\documentclass[journal]{IEEEtran}
\usepackage{amsmath,amsfonts}
\usepackage{natbib}
\usepackage{cleveref}
\usepackage{xcolor}
\usepackage{csquotes}
  
\usepackage{amssymb}
\usepackage{mathtools}
  \usepackage{widetext}   
    \usepackage{multicol}   
  
 \newcommand{\comment}[1]{}   

\newcommand{\mktall}[1]{}
\renewcommand{\mktall}[1]{\mbox{\rule[-1ex]{0ex}{2ex}{#1}}}

\newcommand{\mypar}[1]{ \qquad  \\ \noindent{\bf #1}}

\def\sign{\mathop{\rm sign}\nolimits}
\def\det{\mathop{\rm det}\nolimits}
\def\tr{\mathop{\rm tr}\nolimits}
\def\argmin{\mathop{\rm argmin}\nolimits}

\def\ArgMin{{\tt ArgMin\;}}

\newcommand{\Eqn}[1]{Eq.\;(\ref{#1})}
\newcommand{\Fig}[1]{Fig.\;(\ref{#1})}

\renewcommand{\Vec}[1]{{\bf #1}}

\newcommand{\SO}[1]{\mathbf{SO}({#1})}

\newcommand{\SLTC}[1]{\mathbf{SL2}({\mathbb{C}})}

\newlength{\ourfigwidthw}
\newlength{\ourfigwidth}
\newlength{\ourfigwidthh}
\newlength{\ourfigwidtht}
\newcommand{\showfile}[1]{}
  \renewcommand{\showfile}[1]{{\tt #1}\\[.2in]}

\newcommand{\ednote}[1]{}
  \renewcommand{\ednote}[1]{{\bf\large{\color{red} Note to Editor:}\ \/}%
       {\color{red} {#1}}}
     \renewcommand{\ednote}[1]{}

\newcommand{\earl}[1]{}
  \renewcommand{\earl}[1]{{\bf\large{\color{red} Note to Editor:}\ \/}%
       {\color{red} {#1}}}

\newcommand{\advanced}[1]{}
\renewcommand{\advanced}[1]{\begin{quote}
  \rule{0.75\ourfigwidth}{.5pt}\\ \nopagebreak
   $\dagger$ {#1}\\ \nopagebreak
   \rule{0.75\ourfigwidth}{.5pt}
   \end{quote}. }

\newcommand{\opt}{\mbox{\small opt}}
\newcommand{\init}{\mbox{\small init}}
 
    \newcommand{\T}{}
 \renewcommand{\T}{\mbox{\footnotesize t}}   
 
 \title{The Determinant Ratio Matrix  Approach to Solving  3D Matching 
 and 2D Orthographic Projection Alignment Tasks}

\author{\parbox{2in}{Andrew J. Hanson\\
Indiana University\\
Bloomington, IN\\  
 {\tt\small hansona@iu.edu} } 
 \parbox{2in}{Sonya M. Hanson\\
The Flatiron Institute\\
New York City, NY\\
{\tt\small shanson@flatironinstitute.org} } }

\begin{document}
 
 \maketitle
 


 
\begin{abstract}    
 Pose estimation is a general problem in computer vision with applications ranging from 
 self-driving cars to biomedical imaging.  The  relative orientation of a 3D reference object 
 can be determined from a 3D rotated version of that
object, or, as is often the case, from a projection of the rotated object to a 2D planar image.  This projection
can be a perspective projection, referred to as the PnP problem, or an orthographic projection, which we
refer to as the OnP problem; in this work, we restrict our attention to the OnP problem and 
the full 3D pose estimation task, which we refer to for convenience as the EnP problem.
  Here we present a novel closed-form solution to both  the EnP and OnP problems, derived by 
applying either the quaternion adjugate or the rotation matrix element  framework to the
 corresponding least squares problem, and 
dubbed the determinant ratio matrix (DRaM) approach.  While our solutions are strictly valid only 
for error-free data, we can adjust noise-distorted improper rotation matrix candidates to their 
nearest exact rotation matrices  with a straightforward rotation correction scheme.  
While the SVD and optimal quaternion eigensystem methods for the EnP task, often 
known as  the RMSD  (root-mean-square-deviation)
problem, determine  the noisy 3D-3D alignment  even more accurately than our method, the noisy 3D-2D 
orthographic  (OnP) task has no known comparable closed form, and is solved  quickly 
and to a high degree of accuracy by our scheme. 
 We note while  previous similar work has been presented in the literature
 exploiting both the  QR decomposition and the  Moore-Penrose pseudoinverse transformations, 
  here we place these methods in a larger context that has not
previously been fully recognized in the absence of the corresponding DRaM solution.   Thus we
term this class of solutions as the DRaM family, and conduct comparisons to the behavior of the
 RMSD family of solutions for the EnP and OnP rotation estimation problems.  Overall, this work
  presents both a new solution to the 3D and 2D orthographic  pose estimation problems and 
  provides valuable insight into these classes of problems.  With hindsight, we can even show 
  that our DRaM solutions to the exact EnP and OnP problems possess derivations that 
  could have been discovered in the time of Gauss, and generalize to all analogous 
  N-dimensional Euclidean pose estimation problems.
  
\end{abstract}


\begin{IEEEkeywords}
 RMSD, OnP,   Alignment, Pose Estimation, Orientation Matching,  Attitude Extraction, Orthographic Projection.
\end{IEEEkeywords}

 
\section{Introduction}
\label{sec:intro}

{\Large F}{inding} the best  rotation aligning a reference object to an observed, rotated instance of the same object  is a problem relevant to areas ranging from machine vision to astronautics to biochemistry. 
Analytical solutions determining the rotation that solves the least-squares problem for exact or noisy
 3D matched point-cloud objects have existed for decades 
 (see, e.g., \citet{Davenport1968,Schonemann-Procrustes1966,Green-matsoln-1952});
however, no comparable solutions have been presented for the rotation aligning a   
3D reference model  to  a corresponding 2D orthographic or perspective projected  image.  Here we present closed-form determinant ratio matrices (DRaM's), expressed transparently  in terms of data cross-covariances, that solve the error-free  3D-3D alignment (Euclidean distance loss or ``EnP'')
and 3D-2D orthographic projection alignment (Orthographic projection loss or ``OnP'')  problems.  
We reveal in addition the unexpected correspondence 
 of at least two other methods applicable to both EnP and OnP  in the literature that are numerically identical to
the DRaM's.  The work we present here is a significant elaboration and  contextualization of the authors' preliminary
developments of this material, building
on \cite{HansonHansonAdjugateArxiv-May2022v1,HansonQuatBook:2024}; in addition, 
an initial application-driven exploration, exploiting only earlier theoretical aspects, was published in \cite{Lin_Hanson_DRaM_2025_ICCV}.  The methods employed here were applied to solve the more challenging
perspective n-point  (PnP)
problem in \cite{HansonQuatBook:2024}, but a DRaM form of that solution evaded discovery;  an
explicit  DRaM  form for the error-free PnP task has recently been obtained, and the details will be 
elaborated elsewhere \citep{HansonDRaMPnPinprep2025}.

Our results include several components with subtle relationships whose context we outline now to assist
the reader.   Our first step, well-known, for example, from \citep{Horn1987},  is to pose the  EnP and OnP
 problems (for   the 3D-3D and 3D-2D data, respectively) as   least squares loss expressions to be minimized
over the nine rotation matrix elements $R_{ij}$ or the quaternion 4-vectors $q_{i}$, alternatively expressable in 
terms of the ten quadratic \emph{adjugate quaternions} $q_{ij}=q_{i} q_{j}$.  The latter convert the challenging least squares quaternion loss optimizations   to an approachable quadratic form.  The rotation matrix and adjugate 
quaternion matrix  loss-minimizing expressions can each be solved, for error-free data, to give the DRaM solution 
for the pose.  We remark that we came late to the realization of this equivalence, which does not appear in 
 \cite{HansonHansonAdjugateArxiv-May2022v1,HansonQuatBook:2024}.  The first remarkable side-result
 given at the end of Section  \ref{sec:TheDRaMForm} 
 is that the resulting DRaM rotation matrix solutions can be generalized
 to \emph{any  Euclidean dimension}, essentially robbing the 3D quaternion approach of any unique
 properties it might have for EnP and OnP.   What is therefore important is that 
 Section  \ref{sec:TheDRaMForm} 
establishes the existence, via multiple derivations, of a universal ``DRaM class'' of solutions to the
EnP and OnP alignment problems in any dimension for error-free data.
  
  The second important revelation that we present is that these solutions are in fact not all that new,
  but provide  elegant and unexpected  connections to  entire fields of prior methods. 
  Specifically, in Section \ref{sec:DRaM-Class}, we show that the DRaM class has exactly the same
 numerical properties (and presumably hidden algebraic properties) as the QR-based matrices 
 in the work of  \citet{StegerOrthoOnP2018} and the distinct pseudoinverse reference data  mapping methods
 of,  e.g.,  \citet{DementonDavisIJCV1995} or \citet{HajderWPnPvisapp2017}.
    
The introduction of measurement error in the data deforms the DRaM family of
 candidate rotation matrices so they are no longer pure rotations.  Only the 3D EnP problem
 has a known algebraic solution for noisy data (the quaternion eigenvalue method and equivalents),  
 but  we propose exploiting  \emph{rotation-correction}  methods  that yield 
 \emph{the closest pure rotation matrix}  to the  deformed DRaM-class of solutions
 (see, e.g.,  \citet{Bjorck-ortho-est-SIAM1971,Higham-ortho-est-SIAM1986,BarItzhack2000}).
   After this corrective procedure,  the DRaM-related methods  are only slightly less accurate
than the  gold standard \ArgMin\  (e.g., Levenberg-Marquardt)  numerical 
 search methods  and can be  somewhat faster.   The complete spectrum of properties for all
 known methods for attacking the EnP and OnP alignment tasks for perfect and noisy data
 are summarized in Table \ref{dataTable.tbl};  we know of no equivalent analysis in the literature.
 Appendix A outlines the original  quaternion-based  derivation  of the DRaM formula that led
 over a period of years to the broader analysis we present here, while Appendix B
 supplies a family of applicable software algorithms in Mathematica, most of which are algebraic
 formulas easily (if less transparently) realized in other languages.


\section{Previous Work}
\label{sec:previous}

Orientation matching has been the subject of  a wide spectrum of publications over the
years.   This task was not a major topic of investigation until fairly recently in human
history, when technological requirements like satellite tracking along with automated data
processing and analysis brought the pose estimation problem and methods of
solving it into focus.  The classic problem of aligning a model's orientation, represented
as a set of 3D points in space, with a measured set of corresponding points in a different
orientation is what we label as the EnP task; this alignment problem
 arose in the early days of astronautics, where it is known 
as ``Wahba's Problem'' \citep {Wahba1965}, also  
referred to as the Orthogonal Procrustes or RMSD (root mean square deviation) problem.
As  the digital age matured, 
 applications in disciplines such as
machine vision and robotics, as well as life sciences   areas such as proteomics, molecular geometry, and electron microscopy, developed requirements in the pose matching problem domain.  Three principal approaches were found to solve the orientation discovery or pose determination tasks.   Several instances of a basic linear algebra method  
(the unitary-symmetric ``polar decomposition''  theorem)
came to light independently,  including \citet{Green-matsoln-1952,ThompsonSVD1958},
followed later by   Horn, Hilden,  and Negahdaripour (\citeyear{HornHN1988});  
we will refer to it as ``HHN'' corresponding to the latter frequently-cited  paper.   
 Other studies employing this method include
\citet{GiardinaAES1975,BarItzhackAES1975,SarabandiToR2020} and note that it
 can be proven  to  solve the Frobenius norm minimization, and can be discovered from
 an SVD formulation. 
Soon after Wahba defined the problem,  \cite{Davenport1968} in the aeronautics community
proposed  a quaternion eigensystem solution  (refined in another form by  \citet{Kabsch1976,KabschErratum1978}, and as a higher performance quaternion system in \citet{ShusterQUESTalgo1981}).  

The quaternion eigensystem  approach was rediscovered repeatedly in 
the machine vision and photogrammetry literature  \citep[e.g.,][]{SANSOPhotoGr1973,FaugerasHebert1983, FaugerasHebert1986,Horn1987}, and in the bioinformatics literature 
\citep[e.g.,][]{Kabsch1976,KabschErratum1978,RDiamond1988,
Kearsley1989,Kneller1991,Theobald2005,TheobaldAgrafiotisLiu2010}.  There are two distinct variants,
one a rigorous least-squares approach, usually credited to \citet{FaugerasHebert1983}, that we will
label ``QMIN'', and a more common version based on maximizing the least-squares cross-term, ``QMAX.''
 In parallel, the singular-value-decomposition (``SVD'')  approach of 
\citet{ThompsonSVD1958} reappeared, for example, in  \citet{SchutSVD1960,Schonemann-Procrustes1966,ArunHuangBlostein1987,MarkleySVD1988},
where the standard mathematical source seems to be that given in \citet{Golub-vanLoan-MatrixComp}, Section 12.4, which cites \citet{Schonemann-Procrustes1966}, though a parallel treatment 
appears later in the same journal \citep{CliffSVD1966}. 

 The possible puzzle of the alternative quaternion eigensystem and SVD approaches was resolved, e.g.,  by \citet{CoutsiasSeokDill2004}, who explicitly proved their equivalence.  
 An  alternative  dynamical spring-based system  method has been suggested
by \citet{Doran_etal_PoseEst_ICCV2021}.   We observe that  the quaternion eigensystem method is applicable
only in 3D or 4D \citep[see, e.g.][]{Hanson:ib5072}, while the SVD and HHN methods can be used for data in any Euclidean dimension.   All of these EnP pose discovery methods,
QMIN, QMAX, SVD, and HHN,  are equivalent and  form a single class that will refer to  
as the ``RMSD class'',  in deference to traditional literature;   all recover the same 
``gold standard''  numerical least squares optimal 
rotation (\ArgMin\!)  when applied to \emph{either}  exact or noisy data.
   Further discussion of these solutions to the ENP pose discovery problem can be found, e.g., in
 \citet{Flower1999,forstner1999photogrammetric,Hanson:ib5072,HansonQuatBook:2024}.

The task of pose determination when the only target data are an \emph{orthographic projection}
of the rotated reference data, the OnP problem, is more challenging than one might expect.
While the exact-data case can be solved immediately by several methods, and the optimal
rotation correction problem can be solved by rectangular SVD or the equivalent Bar-Itzhack 
quaternion eigenvalue method, solving the noisy-data OnP  case exactly appears quite challenging. 
The basic requirements of the  OnP problem appear in the context of various approaches to the
 PnP (Perspective n-Point) problem, using terminology such as scaled orthography or
 weak perspective.    
 Various   treatments   have appeared in the literature,  with 
a numerical OnP solution as an 11$^{th}$-degree polynomial given by  \citet{HajderWPnPvisapp2017,HajderOrthoCVICG2019},
and an extensive evaluation of available  techniques  presented 
by \citet{StegerOrthoOnP2018},  who introduces the QR decomposition method.
  The alternative Moore-Penrose pseudoinverse approach to the exact OnP problem 
  is exploited in the  POSIT work of
 \citet{DementonDavisIJCV1995}  and is advocated in \citet{HajderOrthoCVICG2019}.
 While the pseudoinverse system appears  closely related to the QR method, it
 is  distinct in its implementation.   Nevertheless, to the best of our knowledge, while
closed-form noisy-data solutions exist for the EnP problem, no analogous closed-form
solutions have appeared in the literature for the  noisy-data  OnP problem.

Two other closely related problems are significant for our discussion.  One is the
\emph{extraction of a quaternion from a rotation matrix}, which is addressed in
\citet{BarItzhack2000} using an adaptation of the quaternion eigensystem EnP method,
   and the other  is the extraction of the \emph{nearest possible pure rotation}
    from a measurement-contaminated
deformed rotation matrix  candidate, which is given, for example, using an SVD method in 
\citet{Bjorck-ortho-est-SIAM1971,Higham-ortho-est-SIAM1986} as well with an alternative
quaternion  approach in \citet{BarItzhack2000}.   The classic brute force approach to the
quaternion-from-rotation task, checking carefully for the variety of possible singularities, is that
of \citet{Shepperd1978};  the final sections of \citet{SarabandiR2Q2018} include a succinct
rephrasing of the principles of Shepperd's singularity-removal method, which is also
interpretable in terms of the quaternion adjugate-matrix method
 (\citet{LinHansonx2ICCV2023,HansonQuatBook:2024}).
Our preferred   approach  is that of \citet{BarItzhack2000},
based on a reworking of the quaternion pose-extraction method (e.g., \citet{Horn1987}), with
the added observation that it also works with only a $2\times 3$ projection matrix, which in 
fact is also true for the SVD approach.   A contemporary
contribution to the same problem using the polar-decomposition and pointing out its 
relation to the SVD is given by \citet{SarabandiToR2020}.  Four-dimensional quaternion-pair 
 extraction from an $\SO{4}$ rotation was solved  by  \citet{Hanson:ib5072}, and also studied by
 \citet{SarabandiThomas-nearest4D-June2022}.


\section{Fundamental Background and  Methods}
\label{sec:Fundamentals}

Our main narrative takes advantage of
concepts that may be unfamiliar to some readers.  These elements are
essential to establishing the terminology used in the text, and thus we
present them immediately in this section rather than
relegating them to appendices or supplementary material.  
 The topics appearing next include quaternion methods, and particularly
the linear algebra adjugate application and how quaternion adjugate variables
emerge, along with the basic family of ways to define the EnP (3D:3D matching)
and OnP (3D:2D-orthographic matching) pose discovery problems via loss minimization.

\subsection{Rotations in Terms of Quaternions and Quaternion Adjugate Variables}

 A \emph{quaternion} is a point on a 3-sphere  representable as a unit 4-vector  
 $q = [q_0,q_1,q_2,q_3]$ with $q \cdot q = 1$.  Quaternions obey a particular
 algebra, whose details we can ignore here \citep[see, e.g.][for more details]
 {HansonQuatBook:2006,HansonQuatBook:2024},  that allows us to encode any proper 3D 
 rotation using  the following quadratic quaternion form:\\[-.15in]
 \begin{align}
   \MoveEqLeft { R(q) = }  \nonumber & \\ 
    &   \hspace*{-0.3in}
     \begin{scriptsize} \left[ 
\begin{array}{c@{\hspace{-.0 in}}c@{\hspace{-.0 in}}c} 
 {q_0}^2+{q_1}^2-{q_2}^2 - {q_3}^2 & 2 q_1 q_2  -2 q_0 q_3 
& 2 q_1 q_3 +2 q_0 q_2   \\
 2 q_1 q_2 + 2 q_0 q_3  &  {q_0}^2-{q_1}^2 + {q_2}^2 - {q_3}^2   
      &  2 q_2 q_3 - 2 q_0 q_1  \\
 2 q_1 q_3 - 2 q_0 q_2  &  2 q_2 q_3 + 2 q_0 q_1 
 & {q_0}^2 - {q_1}^2 - {q_2}^2 + {q_3}^2
\end{array}  \right]   \end{scriptsize}   .
\label{RotMat.eq}  
\end{align}
 With the single constraint $q \cdot q = 1$, $R(q)$ can easily be shown to
be orthonormal with unit determinant; furthermore, it is clear that any 
rotation optimization task can be formulated exclusively using the quaternion,
 sidestepping the  six orthonormality
constraints on the nine rotation matrix elements by merging them
into the single quaternion constraint.

The \emph{quaternion adjugate variables} are obtained from the 10 unique 
   elements of the symmetric quadratic   quaternion matrix with elements 
   $q_{ij} = q_{i}q_{j}$:
   \begin{equation} \label{AdjMat.eq}
\mbox{Adj}(q_{ij}) =   \left[ \begin{array}{cccc}
       q_{00} &q_{01}&q_{02} &q_{03}  \\
       q_{01} &q_{11} & q_{12}  & q_{13} \\
       q_{02} &q_{12} & q_{22}  & q_{23} \\
       q_{03} &q_{13} & q_{23}  & q_{33} \\
      \end{array} \right]  \ .
     \end{equation}  
     As argued in 
     \citet{HansonHansonAdjugateArxiv-May2022v1,LinHansonx2ICCV2023,HansonQuatBook:2024},
     the form in \Eqn{AdjMat.eq} follows directly from solving the quaternion eigensystem of the
     EnP (also ``Procrustes'' or ``RMSD'') cloud-to-cloud matching problem and expressing the charts of
     the manifold covering the quaternion solution using the adjugate of the corresponding
     characteristic  equation.   Each row or column of the matrix $\mbox{Adj}(q_{ij})$  is  
     equal to a full quaternion vector $q$ multiplied by one of the other four quaternions. 
     We emphasize the fact that there are 14 possible combinations of zeros of $q_i$ rendering
     one or more columns  of \Eqn{AdjMat.eq} unnormalizable, so employing the full  \Eqn{AdjMat.eq} is
     essential for expressing quaternion output in machine learning tasks to avoid the
     14 singular cases occurring inevitably if only a single quaternion 4-vector is used.
     
        We see from \Eqn{RotMat.eq}
     that an alternative  expression for any rotation matrix is therefore
     \begin{align}  \label{qadjFullRot.eq}
    \MoveEqLeft{ R(q_{ij}) =   }    &\nonumber          \\
   &\hspace{-.3in}
   \begin{scriptsize} \left[ \!\! \!
 \begin{array}{c@{\hspace{-1ex}} c @{\hspace{-1ex}} c}
  q_{00} +q_{11}-q_{22} - q_{33} & 2 q_{12}  - 2 q_{03}  & 2 q_{13} + 2 q_{02}   \\
 2 q_{12} + 2 q_{03}  & q_{00} -q_{11} + q_{22} - q_{33}  &  2 q_{23} - 2 q_{01}  \\
   2 q_{13}   - 2 q_{02}     &  2 q_{23}  + 2 q_{01}  &  q_{00} -q_{11} - q_{22} + q_{33} \\
\end{array}  \!\!  \!\right]  \end{scriptsize}   .
\end{align}
          The standard unit-length constraint  $q \cdot q =1$   imposes  a more
           complicated set of \emph{seven} constraints on the adjugate variables,  namely\par
           \vspace{ -0.3in}
\begin{equation} \label{TheqqConstraints.Eq} 
   \!\!\!\! \left. \begin{array}{r @{\hspace{-1.2ex}}c@{\hspace{-1.2ex}} l}
  & q_{00} + q_{11} + q_{22} +  q_{33}  =  1 &\\[0.05in]
  q_{00} \; q_{11} =  {q_{01}}^2 & 
   q_{00} \; q_{22} =  {q_{02}}^2  &
    q_{00} \;  q_{33} =  {q_{03}}^2  \\
    q_{22} \;  q_{33} =  {q_{23}}^2  &
     q_{11} \; q_{33} =  {q_{13}}^2  &
  q_{11} \; q_{22} =     {q_{12}}^2 \\
   \end{array}  \!\!\ \!  \right\}  \! .  
   \end{equation}
   In a moment, we will show how to exploit the quaternion adjugate variables to reduce
  the degree of quaternion-based pose-discovery loss functions by a factor of two, thus
  enabling us to obtain previously unknown explicit algebraic expressions for
  pose-rotation solutions directly   from the least-squares loss functions.
    This is a significant  step beyond the application 
  of the quaternion adjugate variables to machine-learning protocols 
  noted in   \citet{LinHansonx2ICCV2023}.

\subsection{Elements of Quaternion Eigensystems}
\label{sec:CharMatAdjMat }

We will frequently refer to  basic linear algebra methods that have been a mainstay
of the quaternion-based EnP solutions, and which embed the origin of the quaternion adjugate
variables.  The essence starts with a $4\times 4$ matrix $M$, one of whose four eigenvalues, say
$\epsilon_{\opt}$,  has an eigenvector, say $q_{\opt}$, that we need.  The eigenvalues are
computed by  taking the characteristic equation,
\begin{align}  \chi &= \left[ M - \epsilon I_{4} \right]  , \label{chiEqn.eq}  \end{align}
 solving the quartic polynomial $\det \chi(\epsilon) = 0$, and selecting one of the roots
as $\epsilon_{\opt}$.  Classic linear algebra now leads us to an elegant calculation of the
eigenvector corresponding to any single root  $\epsilon_{\opt}$.   Obviously
$\chi( \epsilon_{\opt})$ is singular since $\det \chi(\epsilon_{\opt}) = 0$.   However,
the first step of computing the inverse of $\chi(\epsilon_{\opt})$ is \emph{still legal}:
that is,   the transpose cofactor or \emph{adjugate} of the $4\times 4$ matrix  $\chi$, 
$A_{\opt}  = \text{ Adjugate}\left[\chi(\epsilon_{\opt})\right]$  is singularity-free.
We next exploit the fact that,  by definition, the adjugate construction yields the determinant:
\[  \chi(\epsilon_{\opt})\cdot A_{\opt}   = \det \chi (\epsilon_{\opt})\ . \]
But since $\det \chi(\epsilon_{\opt}) = 0$, we can expand $\chi$ in terms of its
original components \Eqn{chiEqn.eq}, resulting in
\begin{align} 
  \chi  \cdot A & = \left[M - \epsilon_{\opt} I_{4} \right]  \cdot A   =\det \chi(\epsilon_{\opt})I_{4} =  0 \nonumber \\
 &\Rightarrow \  M  \cdot A_{i} = \epsilon_{\opt} \cdot A_{i} \ .
 \end{align}
 Each of the adjugate matrix $A$'s  4-vector columns  $A_{i}$ is therefore proportional to the \emph{same}  
 quaternion eigenvector of $M$.  Some of those  may be proportional to zero, eliminating those columns from consideration; in fact there \emph{fourteen} different possible unnormalizable cases, but since
the eigenvector $q_{\opt}$ obeys the constraint $q\cdot q = 1$, there is always at least \emph{one}
normalizable candidate in the four columns of $A$, and we simply normalize one of those
to always obtain a legal quaternion eigenvector.
For  quaternion-defining matrices $M$, this adjugate matrix is exactly proportional to
\Eqn{AdjMat.eq}, hence the origin of our ``quaternion adjugate variable'' terminology.

 \subsection{ Least-Squares   Loss Measures  for EnP and OnP}
\label{LSQsolns.sec}

For the 3D-3D (EnP) and 3D-2D:Orthographic (OnP) pose discovery problems, exact or
noisy, we will  define least squares loss expressions based upon sets of points in Euclidean space.
Our fundamental data will be a $3\times K$ matrix $\Vec{X} = \{\Vec{x}_{k}\}$ denoting
our model point set, the \emph{reference data}. The 3D points $\Vec{x}_{k}$ are 
arranged in $\Vec{X}$ so that an arbitrary orthonormal proper rotation $R$ in $\SO{3}$ can rotate
from the left to produce the (hypothetically experimentally measured) \emph{target data},
$\Vec{Y} =R \cdot \Vec{X} = \{R\cdot \Vec{x}_{k} \} = \{\Vec{y}_{k}\}$.  For orthographically
projected data, we use only the $2 \times 3$ projection matrix $P = [R_{1},R_{2}]^{\T}$,
and write the image 2-vector data as $\Vec{U} = P \cdot \Vec{X} = 
\{P\cdot \Vec{x}_{k} \} = \{\Vec{u}_{k}\}$.  
We will consider $\Vec{Y}$ and $\Vec{U}$ both with exact data and with Gaussian noise
 with standard deviation $\sigma$ added to the elements of each target vector after
 rotation and possible projection.

  Our basic EnP  loss expression
constructs a sum of 3D  Euclidean squared differences between each reference point and
its matching,  possibly noisy, target point; 2D  Euclidean squared distances between ideal projected
reference points and the target  projected image points are summed  in the OnP variant.  
The sums of Euclidean differences vary smoothly with changes
in the proposed quaternion-based rotation transformation $R(q)$, and the objective
is to find the optimal values $q_{\opt}$ or $R_{\opt} \equiv R(q_{\opt})$ with the
smallest possible loss.  The EnP and OnP losses then are written
\begin{equation}
     \label{3D3DPoseLSQ.eq}  
 \mathbf{S}_{\mbox{\small 3D-3D  EnP}} =
 \sum_{k=1}^{K} \| R(q) \cdot \Vec{x}_{k} -\Vec{y}_{k} \| ^{2}\ ,
 \end{equation} 
 \begin{equation}
     \label{3D2DPoseLSQ.eq}  
 \mathbf{S}_{\mbox{\small 3D-2D Ortho OnP}} =
 \sum_{k=1}^{K} \| P(q) \cdot \Vec{x}_{k} -\Vec{u}_{k} \| ^{2}\ .
 \end{equation} 
 Closed form algebraic solutions are well known for rotations that minimize  the EnP
 \Eqn{3D3DPoseLSQ.eq} with
 exact or noisy data (see, e.g., \citet{Hanson:ib5072} for a review), while only exact
 data, not noisy data, can be accommodated in closed form with solutions known
 at this time for the OnP \Eqn{3D2DPoseLSQ.eq} 
  (see, e.g., \citet{StegerOrthoOnP2018,HajderWPnPvisapp2017}).


 \section{The Determinant Ratio Matrix} 
 \label{sec:TheDRaMForm}
 
  We now present the DRaM formulas solving the exact-data EnP and OnP problems
  and outline their origins and properties.  A synopsis of the authors' original derivation 
  \citep{HansonHansonAdjugateArxiv-May2022v1} that may be of interest
  to some readers is given in Appendix A.     
  Appendix B supplies the basic algorithms for all of the following.
  We begin by noting
 that one can determine from \Eqn{3D3DPoseLSQ.eq} and \Eqn{3D2DPoseLSQ.eq}
 that, when the loss functions are expanded, there are no other variables upon which
 the rotation matrix can depend except the set of the cross-covariance
 elements that can be constructed from the data matrices $\Vec{X} = \{[x_{k},y_{k},z_{k}]\}$,
 $\Vec{Y} = \{[u_{k},v_{k},w_{k}]\}$, and $\Vec{U} = \{[u_{k},v_{k}]\}$.  We denote the
 members of this set as  $\{\text{xx},\text{xy}, \text{xz}, \text{yy}, \text{yz},\text{zz}\}$  and
 $\{\text{xu},\text{yu}, \text{zu}, \text{xv}, \text{yv},\text{zv} , \text{xw}, \text{yw},\text{zw}\}$,
 where  $\text{xx} = \sum_{k} x_{k} x_{k}$,  $\text{xy}=\sum_{k} x_{k} y_{k} $,
$ \text{xz} =\sum_{k} x_{k} z_{k}$ and $\text{ux} = \sum_{k} u_{k} x_{k}$,  
$\text{uy}=\sum_{k} u_{k}  y_{k} $, $\text{uz}=\sum_{k} u_{k} z_{k}  $, and so on.
The essence of the derivation is to  reparameterize the rotation matrices in 
\Eqn{3D3DPoseLSQ.eq} and \Eqn{3D2DPoseLSQ.eq}  as $R(q_{ij})$ 
using the adjugate quaternion variables as shown in \Eqn{qadjFullRot.eq}.
Expanding the resulting loss functions in terms of the $q_{ij}$ and the cross-covariance
sums $\{\text{xx},\text{xy}, \text{xz}, \ldots\}$ results in a pair of complicated but
tractable expressions quadratic in $q_{ij}$, much simpler than the original equations
expressed quartically in $q_{i}$.  
  In summary, after trying many combinations of
the   loss function derivatives with respect to the ten $q_{ij}$ and the seven
constraints  \Eqn{TheqqConstraints.Eq} using the Mathematica solver,
two distinct combinations were found to yield exact-data solutions for the EnP and the OnP
system in 0.5 seconds and 7.0 seconds, respectively (details are given in Appendix  A).
 Imposing \emph{all}  derivative equations and constraints together would yield
 a solution that would solve the least squares problem with or without error, but
 we could not find such a solution in DRaM form.   The exact-data solutions in fact  can be 
 extended easily from the quaternion-restricted 3D space
domain to the  EnP or OnP problem in \emph{any} ND Euclidean space, 
as we will show explicitly below.

\mypar{The EnP Case.} When we insert the ten solutions for $\{q_{ij}\}$,
into the nine rotation matrix expressions in  \Eqn{qadjFullRot.eq}, 
we find that the rotation-matrix elements are composed of ratios of simple $3\times 3$ determinants
of the  15 potentially independent cross-covariances, each with the common denominator
given by the self-covariance determinant 
\begin{align}  \label{BasicDRaMDenominator.eq}
d_{0} = \det \left[ \begin{array}{ccc} \text{xx} & \text{yx} & \text{zx} \\
\text{xy} & \text{yy} & \text{zy} \\  \text{xz} & \text{yz} & \text{zz} \\ \end{array} \right] \ .
\end{align}
 Each of the numerators of the rotation matrix elements $R_{ij}$  takes the form
 of  a $3\times 3$   determinant  $d_{ij}$, where, for example,
\begin{align} \label{BasicDRaMNumerator.eq}
d_{11} = \det \left[ \begin{array}{ccc} \text{yx} & \text{zx} & \text{ux} \\
          \text{yy} & \text{zy} & \text{uy} \\  
          \text{yz} & \text{zz} & \text{uz} \\ \end{array} \right] \ .
\end{align}
The remaining elements of the set $\{d_{ij}\}$ are constructed from
 cyclic permutations, with $[u,v,w] $ in the columns $[d_{i1}, d_{i2}, d_{i3}]$,
respectively, and the first two columns having cyclic ordering of the  elements
$[(y,z),(z,x),(x,y)]$.  The full DRaM matrix for the exact-data EnP pose is then
 \begin{align}
  R(x,y,z;u,v,w)
  =& \left[ \begin{array}{ccc}
 \dfrac{\textstyle d_{11}}{\textstyle d_{0}} &     
                  \dfrac{\textstyle d_{12}}{\textstyle d_{0}}& 
                            \dfrac{\textstyle d_{13}}{\textstyle d_{0}}\\[0.15in]
  \dfrac{\textstyle d_{21}}{\textstyle d_{0}} &
        \dfrac{\textstyle d_{22}}{\textstyle d_{0}}&
              \dfrac{\textstyle  d_{23}}{\textstyle  d_{0}}\\[0.15in]
          \dfrac{\textstyle d_{31}}{\textstyle d_{0}} &
        \dfrac{\textstyle d_{32}}{\textstyle d_{0}}&
           \dfrac{\textstyle  d_{33}}{\textstyle  d_{0}}\\ 
          \end{array} \right] \ .
\label{EnPSolnRot.eq}      
 \end{align}   
It is easy to verify for error-free data that $R(x,y,z;u,v,w)$  gives
the exact same numerical answer as     
any other formulas, as  tabulated in  Table \ref{dataTable.tbl}.
However, as one can immediately see,  the DRaM formula has  the remarkable advantage
that one can actually see how the elements of the optimal pose-determining
rotation matrix relate  to the underlying cross-covariance structure and geometry.
If this problem had come to the attention of Gauss and his contemporaries,  
this formula could   have been part of the mathematical folklore for the last two hundred years.

\mypar{The OnP Case.} We originally considered the adjugate quaternion expansion of the OnP
loss function   \Eqn{3D2DPoseLSQ.eq} independently, and by a similar but
distinct solution procedure (see \citet{HansonHansonAdjugateArxiv-May2022v1} and Appendix A),
we found solutions for the ten $q_{ij}$ that were very
complicated expressions of the cross-covariances, involving   square
roots with very long and seemingly intractable arguments.  However, when substituted into the top
two rows of  \Eqn{qadjFullRot.eq}, all the complications collapse, yielding
a  $2\times 3$ projection matrix $P(x,y,z;u,v)$ solving the exact-data  OnP
problem that was \emph{exactly} the top two rows of the EnP solution \Eqn{EnPSolnRot.eq}.
This was what it had to be for exact data, since the bottom row of 
a $3\times 3$ rotation matrix must  be the cross-product of the first two rows
for consistency.  Nevertheless,  the required consistency can
be discovered independently for the OnP problem.  The bottom row itself,
given only the 2D OnP data $\Vec{U} \{[u_{k},v_{k}]\}$, 
happens to have its own elegant form computable from the cross-product as
\begin{equation} \label{OnPPose3rdRowB.eq}
    \tilde{d}_{31}\to \left[ \begin{array}{ccc}
 \text{xx} & \text{ux} & \text{vx} \\
 \text{xy} & \text{uy} & \text{vy} \\
 \text{xz} & \text{uz} & \text{vz} \\
\end{array} \right]   .  \\
  \end{equation}
If we write \Eqn{OnPPose3rdRowB.eq} as $\tilde{d}_{31}(x,u,v)$, then the rest of the bottom
row of the OnP exact-data DRaM is $\tilde{d}_{32}(y,u,v)$ and $\tilde{d}_{33}(z,u,v)$, and we
can write the OnP pose solution projection matrix, extended from $2\times 3$ to $3\times 3$, as
\begin{align} 
  P(x,y,z;u,v)  
 = &  \left[ \begin{array}{ccc}
 \displaystyle\frac{\textstyle d_{11}}{\textstyle d_{0}} &     
                  \displaystyle\frac{\textstyle d_{12}}{\textstyle d_{0}}& 
                        \displaystyle    \frac{\textstyle d_{13}}{\textstyle d_{0}}\\[0.15in]
 \displaystyle \frac{\textstyle d_{21}}{\textstyle d_{0}} &
       \displaystyle \frac{\textstyle d_{22}}{\textstyle d_{0}}&
             \displaystyle \frac{\textstyle  d_{23}}{\textstyle  d_{0}}\\[0.15in]
         \displaystyle \frac{\textstyle \tilde{d}_{31}}{\textstyle d_{0}} &
       \displaystyle \frac{\textstyle \tilde{d}_{32}}{\textstyle d_{0}}&
          \displaystyle \frac{\textstyle  \tilde{d}_{33}}{\textstyle  d_{0}}\\ 
          \end{array} \right] \ .
\label{OnPSolnRot.eq}      
\end{align}

 \section{Equivalent DRaM Derivation using Least Squares with Rotation Matrix Itself}
 \label{sec:RotationMatrixDRaMDerivation}

The previous Section describes the quaternion-motivated research process that led to the DRaM solutions
for EnP and OnP in the form of \Eqn{EnPSolnRot.eq}  and \Eqn{OnPSolnRot.eq}.  
However,  \emph{once that form was known},  we began to wonder if we had neglected other
simpler approaches to these forms:  in fact,  we had missed an important alternative, made obvious
by doing a reverse parameterization of \Eqn{qadjFullRot.eq}, which takes the form \par       
\vspace{-.1in}   \begin{scriptsize}  \begin{align}  \label{qadjAsRot.eq}
 \left\{ q_{00} \to \frac{1}{4}\left(r_{11}+r_{22}+r_{33}+1\right),
    q_{11} \to \frac{1}{4}\left(r_{11}-r_{22}-r_{33}+1\right),  \nonumber  \right.\\
     q_{22} \to \frac{1}{4}\left(-r_{11}+r_{22}-r_{33}+1\right),
     q_{33} \to \frac{1}{4}\left(-r_{11}-r_{22}+r_{33}+1\right), \nonumber \\
      q_{01} \to \frac{1}{4} \left(r_{32}-r_{23}\right),
     q_{02} \to \frac{1}{4} \left(r_{13}-r_{31}\right),  
     q_{03} \to \frac{1}{4} \left(r_{21}-r_{12}\right),   \nonumber\\
 \left.     q_{23}\to \frac{1}{4} \left(r_{23}+r_{32}\right),
      q_{13}\to \frac{1}{4} \left(r_{13}+r_{31}\right),
      q_{12} \to \frac{1}{4}\left(r_{12}+r_{21}\right)    \right\} \ .
 \end{align}    \end{scriptsize} 
 Therefore one would expect that simply substituting the matrix
 \begin{align} \label{theRijArray.eq}
 R (r_{ij})& = \left[ \begin{array}{ccc}
    r_{11} & r_{12} & r_{13} \\
    r_{21} & r_{22} & r_{23}\\
    r_{31} & r_{32}  &r_{33}
    \end{array} \right]
  \end{align}
  into the loss Eqs.\ (\ref{3D3DPoseLSQ.eq}) and (\ref{3D2DPoseLSQ.eq}) with some
  constraints could also yield the DRaM formulas.    We had rejected this possibility
  originally because our hypothesis was that we could only solve the least-squares
  problem with  the nine $r_{ij}$  derivatives of the loss functions if we included
  constraints such as the six conditions  $\sum_{k} r_{ik}r_{jk} = \delta_{ij}$  on the rows,
  or at the very least the constraint $\det R=1$, analogous to  the adjugate
  quaternion constraint $q_{00} + q_{11} + q_{22} + q_{33} = 1$.  These symbolic
  optimization problems over the $r_{ij}$ variables
  produced empty results with our  available Mathematica  solution software.
  However, we were aware that, for the perfect-data case, we had in some cases achieved correct
  quaternion-based solutions even with missing constraints.  When we finally applied the
   solver to the loss  functions with \emph{only} the nine vanishing $r_{ij}$ derivatives,
   and  with  \emph{no}  constraints,   both the EnP and OnP systems immediately
    yielded the DRaM   equations.   Theoretically, once the system with full $r_{ij}$
    constraints  failed, that would have removed that method from consideration;
    why eliminating the constraints, which would be necessary for a noisy-data
    solution, gives an unsuspected exact-data solution is puzzling.
    But, in any case,  we report that we can use either set of variables, $\{q_{ij}\}$ or $\{r_{ij}\}$,
    to  produce exact-data DRaM pose solutions for EnP and OnP, provided  one
    stumbles on the serendipitous combination of constraints or lack thereof to
    supply to the solver.    In the next Section, we will exhibit our discovery of 
    a  \emph{third}  unexpected DRaM derivation that is entirely independent of
      any process involving the optimization of loss functions.


\section{Independent DRaM  Derivation and Extension to All Dimensions.}
\label{sec:NDderivation}
  
Once \Eqn{EnPSolnRot.eq} and  \Eqn{OnPSolnRot.eq} present themselves
 as the result of solving the EnP and OnP least squares problems,  one recognizes
 that the  structures of the DRaM numerator determinants have very suggestive
 linear algebraic properties.   One is led to  look at the transformation of the reference 
elements $\Vec{x}_{k} = [x,y,z]_{k}$ to the elements of the target data 
$\Vec{y}_{k} = [u,v,w]_{k}$  explicitly for each $k$, which take the form\\[-0.1in]
\begin{equation}
   \begin{array}{rl}
 u \ = & r_{11} \,x + r_{12} \, y + r_{13}\, z \\
 v \ =  & r_{21}\, x+ r_{22} \, y+ r_{23} \, z \\ 
 w\ = &   r_{31}\,x+ r_{32} \, y + r_{33} \,z  \ .\\  
   \end{array}  
   \label{uvwRot.eq}
   \end{equation}
  Then, since the typical mixed cross-covariances have the form 
\begin{equation}
\text{ux} = \sum_{k} x_{k} u_k = \sum_{k} x_{k}\left( r_{11}x_{k} + r_{12}y_{k} +r_{13}z_{k} \right) \ ,
\end{equation}   
we can look at a representative determinant with one column of mixed cross-covariances 
and write out the reduction of the expression  
 \begin{equation}
\begin{aligned}
        d_{11} & =  \det \left[ \begin{array}{ccc}
 \text{xy} & \text{xz} & \text{ux} \\
 \text{yy} & \text{yz} & \text{uy} \\
 \text{zy} & \text{zz} & \text{uz} \\ \end{array}  \right]\\[0.025in]
    &=   \   \det \left[ \begin{array}{ccc}
 \text{xy} & \text{xz} &  r_{11}\;\text{xx} + r_{12\;}\text{xy} + r_{13}\;\text{xz}  \\
 \text{yy} & \text{yz} &   r_{11}\;\text{xy} + r_{12}\;\text{yy} + r_{13}\;\text{yz} \\
 \text{zy} & \text{zz} &   r_{11}\;\text{xz} + r_{12}\;\text{zy} + r_{13}\;\text{zz} \\ \end{array} \right] \\[0.025in]
  &=   \  \det \left[ \begin{array}{ccc}
 \text{xy} & \text{xz} &  r_{11}\;\text{xx}    \\
 \text{yy} & \text{yz} &   r_{11}\;\text{xy}   \\
 \text{zy} & \text{zz} &   r_{11}\;\text{xz}   \\ \end{array} \right]   = \ \ r_{11}\; d_{0}  \ .  
\end{aligned}
 \label{proofOfR11Match.eq}
 \end{equation} 
 
  \noindent Strikingly, by the standard determinant column-subtraction equivalence rules,
  everything cancels out except the overall multiple of the rotation matrix element $r_{11}$
  by the self-covariance determinant $d_0$, which is exactly the DRaM denominator.
  The same argument holds for all the $d_{ij}$ determinants, 
  and so \Eqn{EnPSolnRot.eq} is identical to the rotation matrix $R = [r_{ij}] $,
  and our proof is complete.  We can
  easily see that for cloud distributions in $N$-dimensional Euclidean space,  the exact
  same proof goes through, and, for error-free data, we can  compute the $\SO{N}$  
  matrix for the EnP task, rotating one $N$-dimensional cloud into another using simple determinants
  of the cross-covariance elements.  Similar arguments hold for the OnP case involving $(N-1)$-dimensional
  orthographic image data, with the
  additional  exploitation of  the $N$-dimensional generalization of the cross-product
  to obtain the last row of the  $\SO{N}$ rotation.   Therefore, the entire EnP and OnP
  DRaM representations solving  the error-free  pose-estimation problem follow from the most elementary
  rules of linear algebra, with no connection to the least-squares optimization problems that
  led us originally to the DRaM.

\qquad


\section{Additional Members of the DRaM Class of Exact-Data Solutions: QR and Moore-Penrose}
\label{sec:DRaM-Class}

Once the DRaM solutions are known, further investigation reveals  that
the DRaM is not an isolated case, but is in fact the most  intuitive form belonging to a \emph{class}
of (at least) three distinct algebraic methods with \emph{identical} numerical behavior under all
conditions of which we are aware.  In particular, all three solve the exact-data cases of both EnP and OnP,
and all produce rotation matrix candidates that deviate from orthonormality in exactly the same
way for all reasonable values of noise and model data sets.  As noted by \citet{HajderWPnPvisapp2017}, 
at least  four data  points are generally required for this class. We now outline the properties of
these alternative methods, assembling detailed comparisons of the assorted techniques  
below in  Table \ref{dataTable.tbl}.

 \mypar{ QR Decomposition Transformation  of Loss Function to Trivial
   Frobenius Norm  of matrices.}  Our first alternative solution of the DRaM class is 
 based on an exploitation of the QR decomposition suggested in the extensive review
 of the OnP problem by \citet{StegerOrthoOnP2018}.  This method takes  the reference data array $\Vec{X}$
 appearing in both Eqs.\ (\ref{3D3DPoseLSQ.eq}, \ref{3D2DPoseLSQ.eq}) and rearranges its
 QR decomposition to convert  the conventional sum of Euclidean squared differences
 in either 3D or 2D to a Frobenius difference between two small matrices, that is,
 to the sum of squared differences of all the elements of these matrices.
 We begin with the  fundamental  QR decomposition, which  is
     \begin{equation}
         \label{QRBasic.eq}  
     \mathbf{QR}\left( \Vec{X} \right) =  \{ S,T \} \text{\large \ \ where  \ } \Vec{X} = S^{\T} \cdot T \ .
    \end{equation} 
 For our default data array dimension $\Vec{X}\rightarrow 3 \times K$,  the QR elements
 are  $S \rightarrow 3 \times 3$  obeying $S^{\T}\cdot S  = I_{3}$ and  $T \rightarrow 3\times N$ 
 upper  triangular in the left three columns.  This enables us to multiply the $3 \times K$ 
 form of the squared loss  expression on the  right by the Moore-Penrose 
 pseudoinverse $T^{+}$ of $T$ to perform the transformation
  \begin{equation}
         \label{QRBasicA.eq}  
         \| R \cdot \Vec{X} - \Vec{Y} \|_{2}^{2} \rightarrow \| R \cdot S^{\T}  -   \Vec{Y}\cdot T^{+} \|_{2}^{2} \ .
         \end{equation}
The exact same equations hold for OnP with $R \rightarrow P$ and the $3 \times K$
target data $\Vec{Y}$ replaced by the $2 \times K$ projected data $\Vec{U}$.
Since for square matrices, the sum of columns-squared is the sum of squares of all elements,
the transformed EnP and OnP losses can then be written not as
a sum of $K$ separate Euclidean vector differences but as much simpler
Frobenius norms of $3\times 3$ matrices for EnP or $2 \times 2$ matrices for OnP:
(\citet{StegerOrthoOnP2018}):
\begin{align}
     \label{3D3DPoseQR.eq}  
 \mathbf{S}_{\mbox{\small  QR EnP}} =&
   \| R(q) \cdot  S^{\T}  - \Vec{Y} \cdot T^{+} \|_{\text{Frob}} ^{2} \\[0.2in]
     \label{3D2DPoseQR.eq}  
 \mathbf{S}_{\mbox{\small QR OnP}} =&
  \| P(q) \cdot S^{\T}  - \Vec{U} \cdot T^{+} \| _{\text{Frob}} ^{2}\ .
 \end{align} 
  However, we now observe that can go \emph{one step further} than the transformation employed 
 in  \cite{StegerOrthoOnP2018},   to obtain  \emph{explicit closed form solutions} for the OnP pose rotation,
 provided the data are exact.  
   The additional transformation simply inserts an identity that reduces each half of the squared loss
   to a multiplication by the \emph{entire} inverse of  the QR map of $\Vec{X}$ as follows:
  \begin{equation}
         \label{QRBasicB.eq}  
         \| R \cdot \Vec{X} - \Vec{Y} \|_{2}^{2} = \| R   -   \Vec{Y}\cdot T^{+}\cdot S \|_{\text{Frob}}^{2} \ .
         \end{equation}
   One can easily verify that  
   \begin{align}
     \label{3D3DPoseQRX.eq}  
 \mathbf{R}_{\mbox{\small  QR EnP opt}} =&
    \Vec{Y} \cdot T^{+}\cdot S  \\
      \label{3D2DPoseQRX.eq}  
 \mathbf{P}_{\mbox{\small QR OnP opt}} =& 
   \Vec{U} \cdot T^{+}\cdot S  \ 
 \end{align}  
  immediately solve the exact-data pose discovery problem for EnP and OnP,
  while the noisy-data rotation matrices have corrupted orthonormality.  
  Thus the QR method requires the same rotation correction that we have
  already noted for the DRaM.     
       
  \qquad 
       
   
   {\it Remark:}  The QR decomposition works slightly differently, avoiding
   the pseudoinverse, and corresponding more closely to the presentation of Steger,
   if one presents the data arrays as $\Vec{X} \rightarrow K\times 3$, i.e.,
   with loss element $(R \cdot \Vec{X}^{\T} - \Vec{Y}^{\T})$. In this case,
   $S \rightarrow 3 \times K$, still satisfying $S\cdot S^{\T} = I_{3}$,  but
   now $T \rightarrow 3 \times 3$ is \emph{square} upper triangular and normally
   admits an ordinary inverse.  The
   multiplications in the loss elements are therefore reversed, so 
   \[ R_{\opt} = \Vec{Y}^{\T} \cdot S^{\T} \cdot \left(T^{\T}\right)^{-1} \ . \]

      \qquad

    \mypar{Moore-Penrose PseudoInverse Map of Reference Array.}   
    Our second alternative solution in the DRaM class involves the Moore-Penrose
    pseudoinverse. We already exploited the pseudoinverse
    in the QR decomposition above, but in fact, as suggested, for example, in the
     POSIT treatment of  \citet{DementonDavisIJCV1995} and in  \citet{HajderOrthoCVICG2019}, 
     we can also carry that idea one step further, omitting the QR decomposition altogether.   Instead
    of applying the pseudoinverse to the $T$ matrix in $\Vec{X} = S^t \cdot T$, we can
    eliminate  $\Vec{X}$ entirely  by multiplying inside the loss expression by the
    pseudoinverse    
    \begin{equation} \label{PseudoInvDef.eq}
        \Vec{X}^{+}  = (\Vec{X}^{\T} \cdot \Vec{X})^{-1}  \cdot \Vec{X}^{\T}
    \end{equation}
    of the data matrix $\Vec{X}$ \emph{itself}, in either $3\times K$
    or $K \times 3$ format.  One can verify that,  given exact
    data for both the EnP loss and the OnP loss, the correct rotation matrices
    are obtained from  
     \begin{align}
     \label{3D3DPosePI.eq}  
 \mathbf{R}_{\mbox{\small  P.I. EnP opt}} =&       \Vec{Y} \cdot \Vec{X}^{+}   \\
       \label{3D2DPosePI.eq}  
 \mathbf{P}_{\mbox{\small P.I. OnP opt}} = &      \Vec{U}\cdot \Vec{X}^{+}   \ .
 \end{align}  
 Written out explicitly, these become
     \[  \mathbf{R}  = \sum_{k}  Y_{3k}\cdot \text{PseudoInverse} [X_{3k}] \  \]
 \[  \mathbf{P} = \sum_{k}  U_{2k} \cdot \text{PseudoInverse} [X_{3k}]  \ .  \]   
  One can recognize  the pseudoinverse solutions as closely related to the 
  result of the  transition from the initial QR decomposition in  \Eqn{3D3DPoseQR.eq} and \Eqn{3D2DPoseQR.eq}
  to \Eqn{3D3DPoseQRX.eq} and \Eqn{3D2DPoseQRX.eq}.  
  
 \newpage
  
  {\noindent \bf Noise Abatement.} 
    For noisy data, the candidate rotation matrices of all of the last three methods, the QR decomposition 
of $\Vec{X}$ and the pseudoinverse of $\Vec{X}$,  as well as the DRaM of cross-covariances, 
deviate from orthonormality in  exactly the same way.  This defect can be  partially  
healed using the rotation correction methods introduced in  Section \ref{sec:BarItzhack}, which are
themselves  based on the algorithms in the next section that produce gold standard orthonormal rotation 
matrices solving the EnP problem   for  data with or without error.
 Even when corrected,  none of the DRaM class of algorithms correspond exactly to the 
 gold standard pose solutions,  though they are very close.


 \section{The RMSD Class: Methods that Exactly Solve EnP with or without Noise}
\label{sec:RMSDsolnClass }

For the EnP 3D cloud alignment task, we know a family of algorithms that produce the optimal
aligning rotation without regard to whether the data are exact or noisy, unlike the DRaM class
we have just presented.  We  refer  to this set of methods as the ``RMSD Class'' for convenience.
Several of these methods supply our rotation correction needs in addition to their roles in the EnP
problem, so it is important to list them all explicitly here.  Remarkably, none of these approaches
(except \ArgMin) can solve the noisy OnP problem \citep{StegerOrthoOnP2018}.  The best we
can do is to replace the target $\Vec{Y}$ data in the EnP problem by a 3D constant-$z$ extension
of the 2D OnP  $\Vec{U}$ data.  This does result in pure uncorrupted orthonormal rotation
matrices, but their properties are very poor.  We now examine these methods in turn; all
give the same numerical answers, despite very distinct appearances, though some 
(for example the 3D SVD method and the maximal quaternion eigensystem method), 
have been proven identical \citep{CoutsiasSeokDill2004}.  As for the DRaM class,  
  the RMSD class methods have their properties given in  Table \ref{dataTable.tbl}.

\mypar{1. ArgMin:}   The ``Gold Standard'' for least squares problems such as EnP and OnP (as well
as PnP) is numerical search in the space of rotations to find the, hopefully global, minimum value
of the loss.   The basic algorithm is typically represented in quaternion space, e.g.,
\begin{align} \label{ArgMinSoln.eq}
 q_{\opt} & =  \raisebox{-.9em}{ $\stackrel{\textstyle \mathop{\argmin}}  {q}$}
      \left\| R(q) \cdot \Vec{X} - \Vec{Y} \right \|^{2}  , 
  \end{align}
subject to the constraint $q\cdot q = 1$, which is generally more stable than searching
with the constraints of the full rotation matrix.  Typically Levenberg-Marquardt algorithms
handle this well, while  the 11$^{th}$-degree polynomial Lagrange-multiplier method
that \citet{HajderWPnPvisapp2017}   proposed for OnP should also be sufficient.
 The solution to the pose problem is then $R_{\opt} = R(q_{\opt})$. 
 It is this value against which all other methods must be compared
for evaluation.  \emph{Using the data-simulation defining rotation $R_{\init}$ is simply wrong.}
Once noise is introduced, $R_{\init}$ is completely
uncomputable, and no possible single algebraic formula  independent of the input data
can ever recover its value.  In contrast, there are
actual algebraic formulas, with explicit examples in this section, that can extract the
exact \ArgMin value of the optimal pose from any data.

{\noindent \bf   2. QMIN:}  The classic full least squares formula  \Eqn{3D3DPoseLSQ.eq} was solved
as a quaternion eigensystem problem requiring the computation of the \emph{minimum}
quaternion eigenvalue by \citet{FaugerasHebert1983}.  By a clever insertion of an identity
quaternion expansion, they reduced \Eqn{3D3DPoseLSQ.eq} to a $4\times 4$ quaternion matrix
system of the form
\begin{equation} \label{RMSD-F-Hebert.eq}
 q \cdot \left[ B(x,y,z;u,v,w) \right] \cdot q \ ,
 \end{equation}
 where the $B$-matrix is defined as follows:   
 first we define, for each $k$, the matrix $A(\Vec{x}_{k},\Vec{u}_{k}))$  as
\begin{equation} \label{Adef.eq}
   A _{k} = \left[ \begin{array}{cccc} 
      0 & -a_1 & -a_2 & -a_3 \\ 
     a_1 & 0 & s_3 & -s_2 \\
     a_2 & -s_3 & 0 & s_1 \\ 
     a_3 & s_2 & -s_1 & 0  \\[0.05in]
    \end{array} \right]_{\textstyle k}
\end{equation}
  where, with ``$a$" for ``antisymmetric''   and   ``$s$" for ``symmetric,'' 
  \begin{align} \label{asDef.eq}
  \left. \begin{aligned}
     a_{\{1,2,3\} }  &= \{ x  - u, \,  y  - v, \, z - w \} \\
     s_{\{1,2,3\} } &=  \{ x + u, \,  y + v, \,  z + w \} 
     \end{aligned}  \right\} \ .  
   \end{align}
Then, for each $k$,  we  convert the $A$ matrix into a symmetric matrix with real eigenvalues,
\begin{align*}
  \MoveEqLeft{B_{k} \, = \, {A_{k}} ^{\T}  \cdot A_{k} }=    \hspace*{  3in} \nonumber   \\
 &  \hspace*{- .26 in} \scriptsize {\left[  
     \begin{array}{@{\hspace{0.01in}}c @{\hspace{+0.04in}}c@{\hspace{+0.04in}}c@{\hspace{+0.04in}}c@{\hspace{0.01in}} } 
    {a_1}^2+{a_2}^2+{a_3}^2 & a_3  s_2 - a_2  s_3 & a_1  s_3 - a_3  s_1 & a_2 s_1 - a_1 s_2 \\
      a_3 s_2-a_2 s_3 & {a_1}^2+{s_2}^2+{s_3}^2 & a_1 a_2-s_1 s_2 & a_1 a_3 - s_1 s_3  \\ 
     a_1 s_3-a_3 s_1 & a_1 a_2-s_1 s_2 & {a_2}^2+{s_1}^2+{s_3}^2 & a_2 a_3-s_2 s_3 \\
    a_2 s_1-a_1 s_2 & a_1 a_3-s_1 s_3 & a_2 a_3-s_2 s_3 & {a_3}^2+{s_1}^2+{s_2}^2 \\ 
      \end{array}  \right]}_{\textstyle k}  & 
\label{Bdef.eq}
\end{align*}
and define our $B$ matrix in  \Eqn{RMSD-F-Hebert.eq} as the sum of these elements:
\begin{equation}\label{sumBk.eq}
B = \sum_{k=1}^{K} B_{k} \ .
\end{equation}
Using the full squared-difference minimization measure \Eqn{3D3DPoseLSQ.eq} requires the global minimal value, so the solution for the optimal quaternion in \Eqn{RMSD-F-Hebert.eq} is the eigenvector of the \emph{minimal} eigenvalue of $B$ in   \Eqn{sumBk.eq}.
If $\epsilon_{\opt}$ is the minimal eigenvalue of $B$, the adjugate of the characteristic equation
$\chi = \left[ B - \epsilon_{\opt}I_{4} \right]$ is proportional to  $\text{Adj}(q)$ in \Eqn{AdjMat.eq}, from
which we can select a nonsingular version of the quaternion eigenvector $q_{\opt}$ determining
$R_{\opt} = R(q_{\opt})$.    This is the approach used by Faugeras and Hebert in the earliest application of the quaternion method 
to scene alignment of which we are aware. \emph{Error-free Eigenvalue:} For exact data,   
$\epsilon_{\opt} \equiv 0 $, independent of any data,  so $q_{\opt}$ can  be calculated immediately from $B$ alone
(which remains data dependent).

\mypar{3. QMAX:}  Perhaps the most common EnP solution is based on the quaternion eigensystem resulting
from maximizing the  negative cross-term of the loss function \Eqn{3D3DPoseLSQ.eq} (see, for example, \citet{Horn1987},
or \citet{Hanson:ib5072} for a review).   The eigensystem then takes the form
\begin{align} \label{HornQmax.eq}
\Delta(R(q);\Vec{X},\Vec{Y}) & = \tr R(q) \cdot E \, = \, q\cdot M(E) \cdot q \ ,
\end{align}
which achieves its optimum when the quaternion $q$ is the eigenvector of the \emph{maximal} eigenvalue of $M(E)$.
Here $E$ is the cross-covariance matrix
\begin{align}\label{Emat.eq}
 E (\Vec{X}, \Vec{Y}) = \Vec{X}\cdot \Vec{Y}^{\T} = \left[ \begin{array}{ccc} \text{ux} &  \text{vx} &  \text{wx} \\ 
      \text{uy} &  \text{vy} &  \text{wy} \\ 
       \text{uz} &  \text{vz} &  \text{wz}\\ \end{array} \right] \ .
 \end{align}
  and  $M(E)$ is the \emph{profile matrix},  a traceless, symmetric  $4\times 4$ matrix 
following from     inserting \Eqn{RotMat.eq} into \Eqn{HornQmax.eq}, and expanding the coefficients to give 
\begin{align}  \label{Mprofile.eq} 
M(E_{ab}) = & \left[ \begin{array}{cc}
  E_{11}+E_{22}+E_{33} & E_{23}-E_{32}   \\
 E_{23}-E_{32} & E_{11}-E_{22}-E_{33}   \\
 E_{31}-E_{13} & E_{12}+E_{21}   \\
 E_{12}-E_{21} & E_{31}+E_{13}  \\
\end{array} \right.   \nonumber \\[0.05in]
 & \left. \hspace{0.2in} \begin{array}{cc}
    E_{31}-E_{13} & E_{12}-E_{21} \\
     E_{12}+E_{21} & E_{31}+E_{13} \\
     -E_{11}+E_{22}-E_{33} & E_{23}+E_{32} \\
  E_{23}+E_{32} &   -E_{11}-E_{22}+E_{33} \\
\end{array} \right] \! .
\end{align}

If $\epsilon_{\opt}$ is the maximal eigenvalue of $M(E)$, the adjugate of the characteristic equation
$\chi = \left[ M - \epsilon_{\opt} I_{4} \right]$ is proportional to  $\text{Adj}(q)_{ij}$ in \Eqn{AdjMat.eq}, from
which we can select a nonsingular version of the quaternion eigenvector $q_{\opt}$ determining
$R_{\opt} = R(q_{\opt})$.   \emph{Error-free Eigenvalue:} For exact data,   
$\epsilon_{\opt} = \tr E_{0}= (\text{xx} +\text{yy}+\text{zz})$  is rotation-independent and $q_{\opt}$ can  be
 calculated immediately from $M$
(which remains rotation dependent), omitting the eigenvalue computation step.

\quad

\mypar{4. SVD:}   The singular value decomposition method (see, for example, \citet{Schonemann-Procrustes1966}  
 among a number of others),  which is known \citep{CoutsiasSeokDill2004} to be identical to the QMAX method,
 follows from examining the SVD of the cross-covariance matrix, $\{U,S,V\} = \text{SVD}\left(E(\Vec{X},\Vec{Y})\right)$.
 One can then express the original matrix as $E(\Vec{X},\Vec{Y}) = U \cdot S \cdot V^{\T}$,   and rearrange
the decomposition elements to form an orthonormal matrix solving the corresponding Frobenius norm
problem, and taking the form
 \begin{align} \label{SVDRopt.eq}
 R_{\opt} &= V \cdot D \cdot U^{\T} \ ,
 \end{align}
 where $D =\text{diag}[1,1,\sign(\det U \det V))$. This version of $R_{\opt}$ is the same as the QMIN and QMAX
 solutions, and can also be applied to a standalone matrix $E^{\T}$  (with no available $\Vec{X}$ and $\Vec{Y}$ data)
 to produce an orthonormal rotation matrix with the smallest possible Frobenius distance from the input matrix $E^{\T}$.
 
 \quad
 
 \mypar{5. HHN:}   Our final example of the error-insensitive RMSD class of solutions to the EnP optimal pose discovery
 problem we will label as ``HHN'' corresponding to the method of \citet{HornHN1988}, although the general
 approach was known some time earlier.  While this method looks very similar to a pseudoinverse method
 (see \Eqn{PseudoInvDef.eq}),   it involves using a matrix square root to construct a 
 perfect rotation matrix from the profile matrix $E(\Vec{X},\Vec{Y})$,  and has two alternative forms:
 \begin{align} \label{HHNRopt.eq}
 R^{1}_{\opt} &= \left( E^{\T}\cdot E  \right) ^{ - 1/2} \cdot E^{\T} \\
 R^{2}_{\opt} &= \left( E^{\T} \cdot E \right) ^{+1/2} \cdot E^{-1}   \ .
 \end{align}
 Similar to the SVD, this method will also produce the closest possible orthonormal rotation to data
 for which only the matrix $E^{\T}$ is given.
 
  \quad

 {\noindent \bf Remark on OnP Adaptations.}  The QMIN and QMAX methods can be applied to OnP $\Vec{U}$ data
if it is extended to 3D by placing it in a plane at some fixed $z$ value, and they guarantee orthonormal matrix
results, but these disagree badly with the ideal \ArgMin values.  The SVD and HHN methods have natural
extensions to rectangular matrices such as $\Vec{X}\cdot \Vec{U}^{\T}$,  but those also have exactly the same
 disagreements with the \ArgMin, as noted in Table \ref{dataTable.tbl}.
   If there is a way to reproduce the exact agreement with \ArgMin for noisy
 OnP data, corresponding with the perfect EnP results, we have not yet discovered it.

\qquad


 \section{Rotation Correction for the DRaM Class of Matrices.}\label{sec:BarItzhack}
       
 The family of pose estimation methods including DRaM, QR, and PseudoInverse
 produces perfect solutions and valid rotation matrices only for exact input data;
 with the introduction of noise, the rotation matrices become corrupted and are
 no longer orthonormal, though they may be fairly close to orthonormal.  In order
 to achieve the best possible information from the DRaM family, we therefore
 are motivated to augment these formulas with a final step, the computation
 of the \emph{closest possible pure rotation} to the output matrices corrupted by noise.
 This can be achieved by repurposing our basic EnP pose estimation methods to
 minimize the Frobenius norm relating a pure rotation matrix to our noisy matrices.
   We thus begin with the loss function
   \begin{align}\label{BIFrobRSLoss.eq}
  S_{\scriptstyle\text{Frobenius}} & =  \|R(q)  - S\|_{\text{Frob}}^{2} =  \text{ [const]} + 
           \text{[const]} \tr\left(R(q) \cdot S^{\T} \right)\nonumber\\
   \rightarrow \ \ \ &  q \cdot M(s_{ab}) \cdot q \ .
   \end{align} 
Here the rearrangement  of the rotation data to produce the $4\times 4$ profile
matrix $M(s_{ab})$ parallels the construction of \Eqn{Mprofile.eq} for the
QMAX pose method, except the matrix indices on $S$ are \emph{transposed} with
respect to those of $E$ in \Eqn{Mprofile.eq}.  Solving this as a quaternion eigenvalue
problem is the method proposed by  \citet{BarItzhack2000}.  However, Bar-Itzhack's
work included several more details: in his ``Version 1'' method, he observes that
the technique works for $2 \times 3$ projection matrices if one simply omits
the bottom row of $s_{ab}$ in \Eqn{BIFrobRSLoss.eq}, while his ``Version 2''
is essentially the QMAX method modified with the intent of discovering the
quaternion of a rotation matrix in a more elegant way than the traditional
method of \citet{Shepperd1978}.  In his ``Version 3'', he notes that, in 
addition to simply computing the quaternion for a $2\times 3$ projection
matrix or a full $3\times 3$ rotation matrix,  the technique is perfectly
well-suited for computing rotation correction, since it minimizes the
Frobenius distance between a noisy rotation and a perfect quaternion-defined
symbolic rotation.


In addition to Bar-Itzhack's adaptation of the QMAX quaternion eigenvalue method
for computing the nearest quaternion to perfect or noisy rotation candidates, each of
the other  members of the class except the QMIN accept a $3\times 3$ or $2\times 3$
matrix as direct input, and, properly arranged, can produce a correction from a noisy
rotation matrix to an orthonormal rotation matrix.  The SVD rotation correction approach
is well known, appearing, for example, in 
   \citet{Bjorck-ortho-est-SIAM1971,Higham-ortho-est-SIAM1986,Schonemann-Procrustes1966}.
   The SVD methods work in any dimension, and are very simple to use, as simply supplying
 the transpose $S^{\T}$  of a candidate rotation $S$ to the SVD method in place of
 the cross-covariance $E$ will generate the corrected rotation as the result of \Eqn{SVDRopt.eq}.
 Alternatively,  one can directly implement the correction algorithm using
   $\{U,S,V\} = \text{SVD}\left(R_{\scriptstyle\text{candidate}}  \right)$.
 One can then express the   orthonormal matrix solving the corresponding Frobenius norm
problem as the transpose of \Eqn{SVDRopt.eq},
 \begin{align} \label{SVDRCorrect.eq}
 R_{\opt} &= U\cdot D \cdot V^{\T} \ .
\end{align}
A similar procedure works for the HNN formulation.
Finally, we note that, just as Bar-Itzhack's method works for abbreviated $2\times 3$
projection-matrix  portions  of a rotation matrix, the SVD  and HNN methods are similarly entirely 
capable of handling rectangular matrices.  For example, if we write 
  $\{U,S,V\} = \text{SVD}\left( \text{Mat23}   \right)$, we find that if we choose the 2-row
  matrix $D_{23} = [[1 0 0], [0 1 0]]$,  then we have a solution for a partial ideal
  rotation matrix of the form
  \begin{align}\label{SVDRCorrect23.eq}
\text{Orthogonal 2 x 3 Matrix} = U \cdot D_{23} \cdot V^{\T} \ .
\end{align}

\qquad

 
\section{Experimental Results}
\label{sec:Experiments}

\begin{figure*}[t!] 
\centering
\includegraphics[trim={0 0 0 0cm},width=7.0 in]{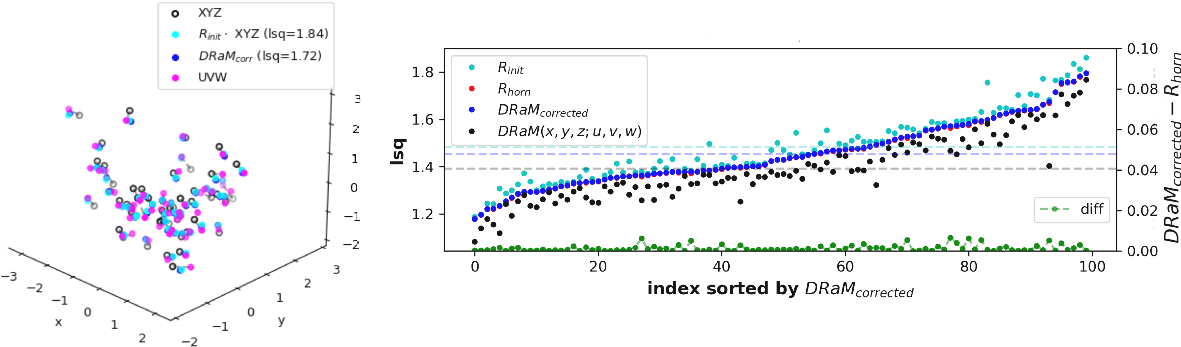}
\caption{Results  for $\sigma = 0.1$, cloud data width  $\sim 2.0$,
using the DRaM analytical solution to the EnP 3D-to-3D  alignment problem.
The  corrected EnP DRaM is not identical to the known-to-be-optimal SVD/Quaternion  ``Horn'' solution,
but it is very close, with 20X scaled differences  plotted at the bottom. \vspace*{ 0.1in}}
\label{3D3DPoseLosses.fig} 
\end{figure*}

 \begin{figure*}[!ht]
\centering
\includegraphics[trim={0 0 0 -1.5cm},width=7.0 in]{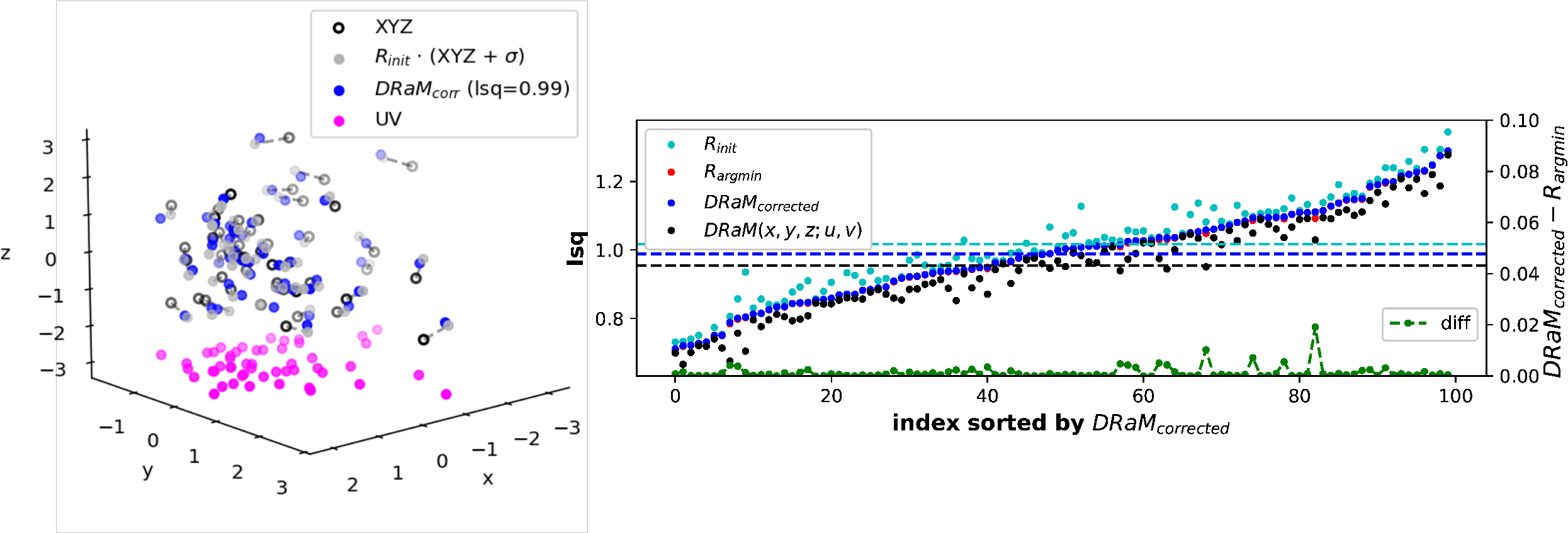}
\caption{Results  for $\sigma = 0.1$, cloud data width  $\sim 2.0$, 
using the DRaM analytical solution to the OnP 3D-to-2D  orthographic projection alignment problem.
The   corrected OnP DRaM is not identical to the known-to-be-optimal  \ArgMin solution,
but it is very close, with 20X scaled differences  plotted at the bottom. \vspace*{  0.1in}} 
\label{OrthoPoseLosses.fig} 
\end{figure*}

We next present some  simple illustrations of the overall quality of our method.
Starting with a list of randomly generated 3D point clouds of unit radius and applying a rotation based on
a uniform randomly selected quaternion, we produced a noise-free alignment problem
either in the form of rotated 3D target cloud, or  a corresponding 2D projection of the
rotated 3D cloud to a plane.   Optional noise with, e.g., $\sigma =0.1$  was added to the
3D target point sets or the 2D target point sets, and each  point set
is recentered with center of mass at the origin to avoid anomalies.
We began by considering three ways of computing a list of  3D-3D and 3D-2D 
losses    \Eqn{3D3DPoseLSQ.eq} and   \Eqn{3D2DPoseLSQ.eq}  for zero noise 
and some typical random noise:\\[0.05in]
\begin{itemize}
\item {  $R_{\init}$:} Compute the loss with the bare rotation used to
  simulate each dataset via $ \Vec{u} = R_{\init} \cdot \Vec{x}+ \Vec{\text err}$,
  truncating  to a 2D plane for the 3D-2D data.
\item { $R_ \text{ \footnotesize DRaM} $:} Evaluate the loss with the DRaM
rotations  \Eqn{EnPSolnRot.eq}, which may be invalid rotations. 
\item { $R_  \text{\footnotesize RotOpt:DRaM} $:} In the case with non-vanishing noise,
correct the warped DRaM rotation candidates to the nearest pure rotation
using a rotation-correction algorithm such as the Bar-Itzhack (denoted ``BI:'') or SVD methods.
\item {$R_{\opt}$:}  Evaluate the loss with the exact optimal aligning rotation matrix
results incorporating added error.   Applying a standard library \ArgMin numerical search
algorithm to \Eqn{3D3DPoseLSQ.eq} and   \Eqn{3D2DPoseLSQ.eq}  is one option.
Exactly the same results  for $R_{\opt}$ are obtained from the (equivalent) SVD
and optimum quaternion eigenvector methods for the 3D-3D EnP problem, with or 
without noise; only the \ArgMin method obtains the true $R_{\opt}$ for the noisy OnP case.
\end{itemize}

For error-free data, all four rotation choices are identical and when 
substituted into the loss functions are uniformly zero to machine
accuracy, consistent with the DRaM matrices
being exact least-squares  loss minimizing solutions to the both pose 
estimation optimization problems.

  \begin{figure*}[!ht]
\centering
\includegraphics[trim={0 0 0 0},width=3.5 in]{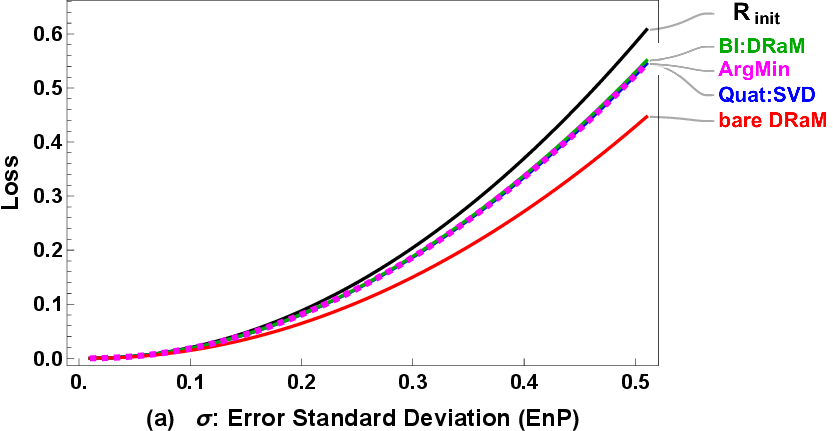}
\includegraphics[trim={0 0 0 0},width=3.5in]{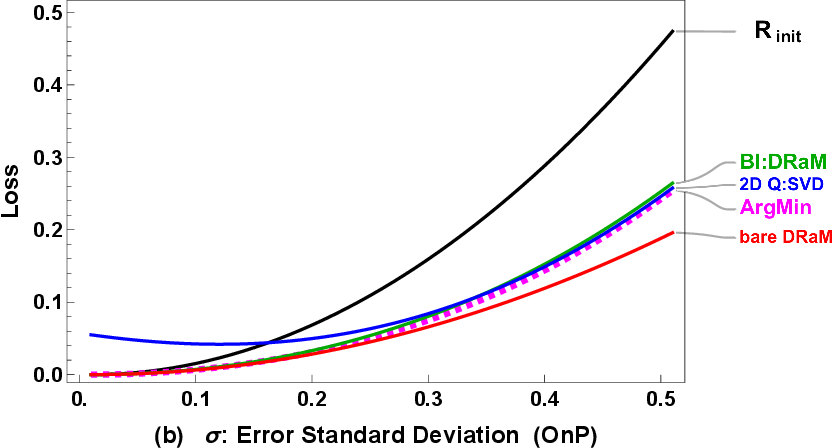}\\
\caption{
(a) EnP with noise.  Behavior of the EnP loss ( \Eqn{3D3DPoseLSQ.eq} ) as a function of standard deviation $\sigma$, data width  $\sim 2.0$,
for a selection of pose estimation methods. 
(b) OnP with noise.  Behavior of the OnP loss ( \Eqn{3D2DPoseLSQ.eq} ) as a function of standard deviation $\sigma$
for a selection of pose estimation methods. Here applying the SVD method with planar target 
data is highly inaccurate at $\sigma=0$ (see Table \ref{dataTable.tbl}), but approaches the other
methods for $sigma > 0.3$.
  For the EnP problem with or without noise, the quaternion-eigenvalue:SVD method  is \emph{exact}, and we see that
 it corresponds to machine accuracy with the \ArgMin gold standard. Note that in both plots, the bare DRaM's loss
is better than the gold standard \ArgMin only because it is deformed and not actually a valid rotation matrix.
}    \vspace*{0.1in}  .
\label{ENPErrorAnglePlotbyMethod.fig} 
\end{figure*}

%
%
   
   \begin{table*}[!ht]   \centering
\begin{tabular}{| l@{\hspace{ 0.02in}} | r  l  |   l || r r    |   l   | }
\hline
\fbox{\large \bf A} & !\!\!Time/$10^5$ & \!\!\!\!\normalsize{loss}\!\!\!\rule[-0.0in]{0in}{4.0ex}&  \parbox{1.6in}{EnP (3D:3D)  Exact Data\\ 
           \hspace*{0.25in} \rule[-0.1in]{0in}{3. ex} $(\Vec{X},\; \Vec{Y} = R_{\init} \cdot \Vec{X} ) $}& 
        Time/$10^5$& \normalsize{loss, angle}  &  \parbox{1.65in}{OnP (3D:2D[Ortho]) Exact Data\\
               \hspace*{0.25in} \rule[-0.1in]{0in}{3. ex} $(\Vec{X}, \;\Vec{U} = P_{\init} \cdot \Vec{X} )$ }\\
          \hline \hline
 \parbox{0.5in}{{\bf\small  ArgMin }  \\[-0.0 in]  {\bf \footnotesize Gold  Std} } &   \rule[-0.15in]{0in}{6.2ex} 
 2755.  & 0& $R _{0} \Leftrightarrow
     \raisebox{-.9em}{ $\stackrel{\textstyle \mathop{\argmin}}  {q}$}
      \left\| R(q) \cdot \Vec{X} - \Vec{Y} \right \|^{2}$      & 
 \rule{0in}{3.0ex}  2054. &  0, \ \ \ 0$^\circ$&   $R _{0} \Leftrightarrow
      \raisebox{-.9em}{ $\stackrel{\textstyle \mathop{\argmin}}  {q}$ }
      \left\|  P(q) \cdot \Vec{X} - \Vec{U} \right \|^{2}$    \\ \hline
{\bf QMIN} &   \rule{0in}{3.0ex}  13.46   & 0  &  $R _{0}  \Leftrightarrow  q_{\mbox{\footnotesize Hebert}}\left( B ( \Vec{X}, \Vec{Y} )  \right) $ &
       13.29 & 0.055, \ \ 37$^\circ$ &  $R _{*}  \Leftrightarrow  q_ {\mbox{\footnotesize Hebert Ortho}}\left( B(\Vec{X}, \Vec{U} ) \right) $ \\
 {\bf QMAX}& \rule{0in}{3.0ex}      8.08   &  0 &  $R _{0}  \Leftrightarrow  q_{\mbox{\footnotesize Horn}}\left( M (E =\Vec{X}\cdot \Vec{Y}^{\T} )  \right) $ &
    7.42&  0.055, \ \ 37$^\circ$ & $R _{*}  \Leftrightarrow  q_ {\mbox{\footnotesize Horn Ortho}}\left( M_{2\times 3} \right) $ \\
 {\bf SVD}& \rule{0in}{3.0ex}  3.06  &  0  &       $R _{0}  \Leftrightarrow   \mbox{SVD}_{3\times 3}\left( E =\Vec{X}\cdot \Vec{Y}^{\T}     \right) $ &
     2.62  &   0.055, \ \ 37$^\circ$&$R _{*}  \Leftrightarrow   \mbox{SVD}_{2\times 3}\left( M_{2\times 3} \right) $ \\
 {\bf HHN}&   \rule{0in}{3.0ex}    2.55& 0 &               $R _{0}  \Leftrightarrow    \mbox{HHN}\left( E =\Vec{X} \cdot \Vec{Y}^{\T}    \right) $ &
     3.62  &   0.055,  \ \ 37$^\circ$ &$R _{*}  \Leftrightarrow   \mbox{HHN}_{2\times 3}\left( M_{2\times 3} \right) $ \\   \hline 
{\bf DRaM}&   \rule{0in}{3.0ex}   5.81 &  0  &  $R _{0}  \Leftrightarrow    \mbox{DRaM}\left( \Vec{X},\, \Vec{Y}    \right) $ &
      4.84 &  0, \ \ 0$^\circ$ & $R _{0}  \Leftrightarrow   \mbox{DRaM}\left(\Vec{X} , \, \Vec{U} \right) $ \\
 {\bf QR}&   \rule{0in}{3.0ex}   1.70    &  0 &   $R _{0}  \Leftrightarrow    \mbox{QR}\left( \Vec{X},\, \Vec{Y}    \right) $ &
      1.34 &  0, \ \ 0$^\circ$ & $R _{0}  \Leftrightarrow   \mbox{QR}\left(\Vec{X} , \, \Vec{U} \right) $ \\
 {\bf PINV}& \rule{0in}{3.0ex}   1.32    &   0 &        $R _{0}  \Leftrightarrow    \mbox{PseudoInverse}\left( \Vec{X}  \right) \cdot   \Vec{Y}   $ &
        1.00 &  0, \ \ 0$^\circ$ & $R _{0}  \Leftrightarrow  \mbox{PseudoInverse}\left(\Vec{X}  \right) \cdot \Vec{U}  $ \\[0.1in]
                   \hline 
                 \end{tabular} 
      
\bigskip

    \noindent       
\begin{tabular}{| l  |   @{\hspace{ .05 in}}r@{\hspace{0.2 in}}r@{\hspace*{.10in}}r @{\hspace*{ .05in}} |  l  || c |  l |}
\hline               
 \fbox{\large \bf B} & \multicolumn{3}{   l |}  {Time/$10^5$   loss \ \ \  angle  \rule{0in}{3.5ex}} &  
     \rule[0.1in]{0in}{3. ex}  \parbox{1.60in}{ EnP (3D:3D) Errorful Data\\ 
       \hspace*{0.25in}  \rule[-0.1in]{0in}{3. ex} $ (\Vec{X},\; \widetilde{\Vec{Y}} = R_{\init} \cdot \Vec{X} + \epsilon ) $} & 
      \! \! \!\!\!Time/$10^5$,\! loss, \!angle \!\! \!\!\!\! \!\!&
   \parbox{1.70in}{OnP (3D:2D[Ortho]) \!Errorful Data\\   
   \hspace*{0.25in}  \rule[-0.1in]{0in}{3. ex} $ (\Vec{X}, \; \widetilde{\Vec{U}}= P_{\init} \cdot \Vec{X} + \epsilon )$ } \\
\hline \hline 

    \rule{0in}{3.5ex} 
\parbox{0.46in}{{\bf\small  ArgMin}  \\  {\bf \footnotesize Gold  Std} } & \rule[-0.15in]{0in}{3.8ex} 2746.& 0.0225&0$^\circ$  &   $R _{1}  \Leftrightarrow   \raisebox{-.9em}{ $\stackrel{\textstyle \mathop{\argmin}}  {q}$} 
  \left\| R(q) \cdot \Vec{X} -\widetilde{\Vec{Y}} \right \|^{2}$  & 
   \rule{0in}{3.0ex}  2097.,  0.0084, 0$^\circ$ & $R _{1}       \Leftrightarrow\!\!  \raisebox{-.9em}{ $\stackrel{\textstyle \mathop{\argmin}}  {q}$ } 
     \left\|  P(q) \cdot \Vec{X} - \widetilde{\Vec{U}} \right\|^{2}$   \\ \hline
  {\bf QMIN} & 13.60& 0.0225& 0$^\circ$ & $R _{1}  \Leftrightarrow  q_{\mbox{\footnotesize Hebert}}\left( B ( \Vec{X},\widetilde{\Vec{Y}} )  \right) $ &
  13.14,  0.042 ,\  31$^\circ$\ \       &  $R _{2 }  \Leftrightarrow  q_ {\mbox{\footnotesize Hebert Ortho}}\left( B (\Vec{X},\widetilde{\Vec{U}} ) \right) $ \\
    {\bf QMAX}    &   8.27& 0.0225 &0$^\circ$ & $R _{1}  \Leftrightarrow  q_{\mbox{\footnotesize Horn}}\left( M (E =\Vec{X}\cdot\widetilde{\Vec{Y}^{\T}} )  \right) $ &
        7.38, 0.042, \ \  31$^\circ$     &  $R _{2}  \Leftrightarrow  q_ {\mbox{\footnotesize Horn Ortho}}\left( M_{2\times 3} \right) $ \\
   {\bf SVD}    &      3.11& 0.0225&0$^\circ$  &     $R _{1}  \Leftrightarrow   \mbox{SVD}_{3\times 3}\left( E =\Vec{X}\cdot \widetilde{\Vec{Y}}^{\T} \right) $ &
    2.66, 0.042, \ \  31$^\circ$    &   $R _{2}  \Leftrightarrow   \mbox{SVD}_{2\times 3}\left( M_{2\times 3} \right) $ \\
 {\bf HHN}    &  2.58& 0.0225& 0$^\circ$  &   $R_{1}  \Leftrightarrow    \mbox{HHN}\left( E =\Vec{X} \cdot \widetilde{\Vec{Y}^{\T} }    \right) $ &
       3.65, 0.042,\ \   31$^\circ$   &   $R _{2}  \Leftrightarrow   \mbox{HHN}_{2\times 3}\left( M_{2\times 3} \right) $ \\    \hline  
{\bf DRaM}    &   5.68 & 0.0227 & 1.42$^\circ$  &  $R _{3}  \Leftrightarrow     \mbox{RotOpt}[   \mbox{DRaM}\left( \Vec{X},\,\widetilde{\Vec{Y}}    \right) $]&
    6.80,  0.0089, 2.85$^\circ$  & $R _{4}  \Leftrightarrow   \mbox{RotOpt}[ \mbox{DRaM}\left( \Vec{X} ,\,\widetilde{\Vec{U}} \right) $]\\
  {\bf QR}   &   1.72& 0.0227&1.42$^\circ$   &  $R _{3}  \Leftrightarrow     \mbox{RotOpt}[   \mbox{QR}\left( \Vec{X},\,\widetilde{\Vec{Y}}    \right) $]&
    3.16,  0.0089, 2.85$^\circ$    &   $R _{4}  \Leftrightarrow   \mbox{RotOpt}[ \mbox{QR}\left( \Vec{X} ,\,\widetilde{\Vec{U}} \right) $]\\
  {\bf PINV}    &     1.33 & 0.0227 &1.42$^\circ$    &   $R _{3}  \Leftrightarrow   \mbox{RotOpt}[  \mbox{PseudoInv}\left( \Vec{X}  \right) \cdot  \widetilde{\Vec{Y}}  $]&
     2.79,  0.0089, 2.85$^\circ$   &     $R _{4}  \Leftrightarrow        \mbox{RotOpt}[ \mbox{PseudoInv}\left(\Vec{X}  \right)\! \cdot\!\widetilde{\Vec{U}} $]\\[0.1in]
                  \hline  
\end{tabular}     
  
  \bigskip
  
\noindent

\begin{tabular}{| l@{\ \ \ }   p{2.8 in}  || @{\ \ \  } p{2.8in} |}
\hline               
  \fbox{\large \bf C}  &     EnP (3D:3D)  Rotation Correction (RotOpt) &    OnP (3D:2D[Ortho])  Rotation Correction  (RotOpt) \\
\hline \hline
   &   \rule{0in}{3.5ex}    $\text{ Bar-Itzhack}  =  \text{QuatToRot[}\, 
     q_{\mbox{\footnotesize Horn}}\left( M (E =R_{\text{approx}}^{\T} ) \right) ]$ &
      $\text{Bar-Itzhack} =  \text{QuatToRot[}\, 
     q_{\mbox{\footnotesize Horn}}\left( M _{2\times 3}(E =P_{\text{approx}}^{\T} ) \right) ] $\\
          
  & \rule{0in}{3.5ex}   $\text{ SVD} =    
     \text{ SVD}_{3\times 3} \left( R_{\text{approx}}^{\T}  \right) $ &
      $\text{ SVD} =    
    \text{ SVD}_{2\times 3}  \left(  P_{\text{approx}}^{\T}  \right) $\\[0.1in] 
         \hline      
\end{tabular}\\ [0.1in]
\caption{ 
   \centerline {Summary of EnP and OnP Pose Discovery Algorithm Properties for Exact and Errorful data.}
 \parbox{6.9in}{  \vspace{0.15in} \rm\small  
In these tables we present eight algorithms that have been used to produce rotation matrices aligning reference point
clouds with observed rotated target data.  The two major categories, EnP and OnP,  are supplied with exact data in
Table {\bf A}, and errorful data with standard deviation $\sigma = 0.1$  in Table {\bf B}.  
Table {\bf C} lists typical rotation correction variants of the algorithms.   We use  a single reference data set 
consisting of  an 8 point random cloud with average radius one unit, and  a $21^{\circ}$ rotation of the target data,
 orthographically projected for the OnP case, with the intent of showing a typical behavior rather than an extensive
 collection of cases.  Times are in seconds for 100,000 repetitions 
 of the calculation in  Mathematica,  on an  Apple M4 Max
 processor,  transforming the input reference and target point arrays to a rotation matrix pose estimation.
 Accuracy is measured by two redundant but useful measures, the actual loss function
 specified in the top \ArgMin entry, and the deviation  from the \ArgMin gold standard angle of 
 each rotation matrix's  angle $\theta$ obtained from  its quaternion $q_{0} = \cos(\theta/2)$.
  The details of the algorithms and programs used are given in the main text and detailed in the
 Supplementary Material.  One should note that  the SVD-like algorithms naturally employ input that is
  transformed from the $n$-dimensional data
 arrays into compact $3\times 3$ or $2\times 3$ cross-covariance matrices, while such data is insufficient for the DRaM
 family of algorithms, which require  the reference data ($\Vec{X}$ in the table) to be supplied separately
 from  the target data ($\Vec{Y}$  and  $\Vec{U}$ in the table).  For OnP, even with exact data,
  the RMSD class  algorithms fail to be exact matches to the \ArgMin gold standards because the 
   corresponding cross-covariances,  which contain all the needed data for EnP, are missing a component of the loss corresponding to non-canceling rotation  matrix terms. } }  
   \vspace{ 0.025 in}
   \label{dataTable.tbl}
 \end{table*}

  In \Fig{3D3DPoseLosses.fig},   we plot
   the values of the  3D-3D  EnP least squares  losses \Eqn{3D3DPoseLSQ.eq}, for all four
   options we considered, 
   sorting by the losses using optimal SVD/Quaternion rotation matrix values, which define 
   the gold standard of comparison to other losses.  A similar plot is shown in 
   \Fig{OrthoPoseLosses.fig} for the 3D-2D OnP orthographic projection losses defined
   by \Eqn{3D2DPoseLSQ.eq} using numerical  \ArgMin  rotation matrix values, which 
   are the gold standard for this case.
      From top to bottom, the worst case is the $R_{\init}$  rotation, whose behavior is
   expected because, as we know from the SVD/Quaternion solution to the noisy 3D-3D
   alignment problem, there exists a closed form algebraic solution (the Cardano formula
   for the quaternion eigenvalues) for $R_{\opt}$, while $R_{\init}$ cannot in any way
   be computed from noisy data.  $R_\text{\footnotesize BI:DRaM}$  is next, plotted
   essentially  on top of the $R_{\opt}$ losses because at this scale they are almost
   indistinguishable:   the $R_\text{\footnotesize BI:DRaM}$ loss
    always  exceeds or equals the $R_{\opt}$ loss, and this amount  magnified by  
    around 20X is plotted on
   the bottom axis.   As we might expect, the unaltered $R_\text{ \footnotesize DRaM}$
   matrix, which is \emph{not a rotation matrix}, but remains a solution to the improper
   least-squares minimization problem, actually appears \emph{below} the loss value
   plot for $R_{\opt}$; this is meaningless because this DRaM matrix is not orthonormal ---
   the smallest possible loss with an orthonormal matrix in the loss function is always
   the $R_{\opt}$ loss.
      
     To address the sensitivity of each method to  error in simulated data,    we present in      
        \Fig{ENPErrorAnglePlotbyMethod.fig}(a,b)    
      some simple studies of the properties of the various pose discovery methods, showing
      the sensitivity of the loss curve for a given pose rotation output to increasing error.
     
     We can see that
      even when the DRaM class of pose extraction algorithms encounters significant
      errors, after rotation correction (noted as ``BI'' for Bar-Itzhack),
      they produce loss responses that are extremely close to the \ArgMin gold
      standard pose estimates, though they are not exact, as are the members of
      the EnP class of   solutions.
      
%

      In summary, depending on the application, the corrected DRaM family candidates for 
the aligning rotation in 3D-3D and 3D-2D can be quite sufficient, and, while they
are not necessarily superior to SVD/quaternion methods for the 3D-3D problem,
they are  much faster to compute for the 3D-2D problem than the available
 \ArgMin  method, which typically  involves a
   Levenberg-Marquardt brute force numerical search of the quaternion rotation space.
   We have selected several properties of each method of the RMSD class and the
   DRaM class for comparison, namely speed, loss measured by the appropriate
   least squares formula, and total rotation difference relative to the \ArgMin gold
   standard solution, and tabulate these all together in the comprehensive  
   Table \ref{dataTable.tbl}  for both exact and moderately noisy data.  The
   results have many interesting features.  For example,
   while typical times for 100,000 OnP pose discoveries are 2,000 seconds for  \ArgMin, for
   the same noisy-data OnP task,  DRaM, QR, and PseudoInverse take about 6 seconds,
   3 seconds, and 2 seconds, respectively.  The numerical results for the three methods
   are identical, so it appears that the PseudoInverse would be the method of choice,
   in agreement with an observation of \citet{HajderOrthoCVICG2019}.
   While the bare DRaM  family encountering data with $\sigma = 0.1$ is quite inaccurate, 
   the result after the Bar-Itzhack-style  correction, with negligible expense,  is typically 
   within 3 degrees of the gold standard.


\section{Conclusion}
   
      The investigation we have described began with the unexpected discovery  that
 the EnP (3D:3D cloud matching) and OnP (3D:2D orthographic cloud matching)  tasks are solvable
 for exact data by determinant-ratio optimal rotation matrices, the DRaMs of  Eqs.\ (\ref{EnPSolnRot.eq})
 and  (\ref{OnPSolnRot.eq}).  Three distinct derivations of these simple algebraic equations
 were introduced, one based on solving  the 3D least squares optimization
 of Eqs.\ (\ref{3D3DPoseLSQ.eq}) and (\ref{3D2DPoseLSQ.eq}) using
 the ten quaternion adjugate $q_{ij}$ variables, the second achieving the same exact result using the
 nine orthogonal 3D rotation matrix variables $r_{ij}$, and the last completely independent of
 the least squares framework and valid not only for 3D space but any N-dimensional Euclidean space
 for both the EnP and OnP tasks.   Further investigation revealed that there is an entire class of
 three algebraic methods with  exactly the same properties as the DRaM function, with this "DRaM
 class" consisting of the DRaM, the QR decomposition mapping applied to the reference data array,
 and Moore-Penrose pseudoinverse applied to the reference data array.  All members of the DRaM
 class produce perfect solutions to the EnP and OnP pose determination problems for perfect
 target data, and all members produce non-orthogonal warped rotation matrices for noisy target
 data, and yet all can be restored to
 highly accurate (but not exactly accurate) orthonormal
 matrices using  rotation correction algorithms.  We also enumerated and
 evaluated a complementary set of four  algorithms,  the  minimal quaternion eigenvalue, the
 maximal quaternion eigenvalue, the SVD, and the HNN methods of the ``RMSD class"
 that produce perfect solutions,   matching the \ArgMin ``gold standard'' value, for all EnP
 problems with either perfect or noisy data.   However, no analogs to the RMSD class algorithms
 are known for the OnP problem, and the  rotation-corrected DRaM class
 algorithms appear to be the best available closed-form methods known at this time.
 
      While we have shown the power of the DRaM class of algorithms,    a number of open questions 
remain.   For example, we know that the   11$^{th}$-degree polynomial OnP solution of 
\citet{HajderWPnPvisapp2017}  is an accurate numerical solution with results equivalent to
the \ArgMin method.   Is it possible that the Hajder solution could reduce to the  sought-for
closed form OnP formula belonging to the RMSD class?   Next, in addition to the DRaM-class
of solutions we have presented for the orthogonal projection  (OnP) problem, the application
of the  DRaM method to the six-degree-of-freedom PnP (perspective projection) problem,
extending the incomplete PnP treatment in \cite{HansonQuatBook:2024}, is known
and will be described elsewhere.
Finally, we can imagine that aspects of the framework we have presented 
here for the pose estimation problem might be effectively  incorporated 
into the machine learning and deep neural network approaches to machine vision.

\section*{Acknowledgments}
This project has evolved over a period of several years, and we benefited from feedback
and suggestions from many colleagues, whose input we thankfully acknowledge.  We are
particularly grateful to B.K.P. Horn, who has suggested readings, provided
guidance, and has always been ready to return an answer to a question.  AJH also thanks David
Crandall for his collegiality and willingness to take the time to help in clarifying some
issues.  We are especially appreciative of several referees who, in rejecting an earlier
version of this work,  provided pointers to exactly the references we needed to go further
and significantly improve our understanding of the problem as whole.  SMH gratefully
acknowledges  the support of The Flatiron Institute, a division of the Simons Foundation.

%
%

{
    
     \bibliographystyle{ieeenat_fullname}
     \bibliography{main}

%

}

 \clearpage

\appendices

     

\section{Details of the Quaternion-Based DRaM Solution} 
\label{DRaM-derived.app}

\ \  \\[-0.4in]

%
  
 
 \noindent
  In this Appendix, we fill in the details of the process of solving the EnP and OnP 
  least squares problems by exploiting the quaternion adjugate form. 
 In   \Eqn{AdjMat.eq}, 
 we defined the 10 quaternion adjugate variables, $q_{ij} = q_{i} q_{j}$,
 whose advantages were introduced in  
 \cite{HansonHansonAdjugateArxiv-May2022v1,LinHansonx2ICCV2023,HansonQuatBook:2024,Lin_Hanson_DRaM_2025_ICCV}.
 Here we review the essential elements, observing, as noted in Section  \ref{sec:TheDRaMForm}, 
 that the quaternion adjugate variables $q_{ij}$ are equivalent to the rotation matrix variables $[R]_{ij} = r_{ij}$,
 but with very different constraint structure. In fact, the following derivations can be duplicated using
 the $r_{ij}$ variables  in  \Eqn{3D3DPoseLSQ.eq}  or \Eqn{3D2DPoseLSQ.eq} 
 with no additionally
  imposed constraints, a fact we had not yet discovered when we developed the derivations in
  this Appendix, which require the use of one or more of the quaternion adjugate constraints.
 As also observed in Section  \ref{sec:TheDRaMForm}, 
   both  the $r_{ij}$ and $q_{ij}$
 derivations using explicit  solutions of the loss-function minimization algebra are essentially irrelevant since,
 with hindsight, the DRaM matrix forms can be proven by hand on the back of an envelope to
 solve the EnP and OnP pose discovery problem for exact data.

We begin the algebraic loss function minimization process with the  observation that if we replace the
rotation in the  3D-3D least squares expression \Eqn{3D3DPoseLSQ.eq} 
by the quaternion adjugate expression $R(q_{ij})$
 shown in  \Eqn{qadjFullRot.eq},    
 the optimization problem becomes
 quadratic in $q_{ij}$ and therefore potentially more easily solvable in terms of
 the cross-covariances of the $\{\Vec{x}_ {k} \}$ and $\{\Vec{y}_ {k} \}$ 
 appearing in  \Eqn{BasicDRaMDenominator.eq} and \Eqn{BasicDRaMNumerator.eq} 
  above.   We remind ourselves that if we could constrain an algebraic solver to impose all seven adjugate
  constraints  \Eqn{TheqqConstraints.Eq},   
  we would have an optimal solution of the
  least squares problems 
    \Eqn{3D3DPoseLSQ.eq} and \Eqn{3D2DPoseLSQ.eq} 
  for both exact and errorful data.  Unfortunately, the equations with 
  all the constraints imposed appears to resist closed-form solution.
  However, if we  write out  the expression  as a quadratic polynomial,\\[-0.2in]
  \begin{align*} \label{qqEnPAlgebra.eq}  
     \MoveEqLeft  {S_{\text{EnP}} (q_{ij}; xx,xy,xz,\ldots;ux,vx,wx,\ldots)}  = &  \nonumber \\
     &  {q_{00}}^{2} f(xx,\ldots)   + \cdots  + q_{00}\; g(xx, \ldots) +   \cdots  \    \ , \\[-0.2in] \nonumber
  \end{align*}  
  and try a variety of combinations of vanishing derivative  conditions
  on  $S_{\text{EnP}} (q_{ij})$ with selections of the constraints  \Eqn{TheqqConstraints.Eq}, 
  we find one that actually works.  Since the constraints imply that the derivatives
  are not independent, and since the only true guarantee of a solution is to require
  \emph{all} the constraints to be enforced, the fact that we found a combination
  that is \emph{perfect} for exact data is perhaps unexpected; if there is a deep
  reason for why this succeeded at all, we have not yet understood it.  The successful
  choice is to treat  vanishing of \emph{all ten} derivatives of $S_{\text{EnP}} (q_{ij})$ with respect
  to $q_{ij}$ as independent conditions, while imposing only the \emph{first} constraint
  in  \Eqn{TheqqConstraints.Eq},  
  namely $ q_{00} + q_{11} + q_{22} +  q_{33}  =  1$.
 Assembling this into the Mathematica constraint-list solver in the form \\
 \vspace{-.1in}
   \begin{equation*} \label{the3DMatchSolver.eq}  
 \begin{aligned}
  & \verb|the3D3DAdjMatchSolns = |  \\
   &\verb|  Module[{eqn = |   S_{ \text{EnP}} (q_{ij})\verb|}, |  \\
   &\verb|   Solve[ {  |\\
   &\verb|     D[eqn, q00] == 0, D[eqn, q11] == 0,|\\
   &\verb|     D[eqn, q22] == 0, D[eqn, q33] == 0,|\\
   &\verb|     D[eqn, q01] == 0, D[eqn, q02] == 0 |\\
   &\verb|     D[eqn, q03] == 0, D[eqn, q23] == 0,|\\
   &\verb|     D[eqn, q13] == 0, D[eqn, q12] == 0 |\\
   &\verb|             q00 + q11 + q22 + q33 == 1},  |\\
   &\verb|    {q00, q11, q22, q33, q01, q02, q03, |\\
   &\verb|                         q23, q13, q12}]].|\\[0.0in]
     \end{aligned}    
\end{equation*}
%
%
returns a complete list of solutions for the $q _{ij}(\text{xx}, \ldots)$ in precisely
  the form of   \Eqn{BasicDRaMDenominator.eq} and \Eqn{BasicDRaMNumerator.eq}, 
  combining into the EnP DRaM expression  \Eqn{EnPSolnRot.eq}  
  when assembled to
  construct the rotation using the components in  \Eqn{qadjFullRot.eq}.  
  The EnP
  solution takes 0.5 seconds to solve and 9  seconds to assemble the $q_{ij}$ into
  a rotation and simplify to the DRaM form.
  As noted in the previous section, any set of error-free data that is regular
  with sufficient degrees of freedom produces \emph{exactly} the correct
  rotation matrix applied to the reference data when the reference and target
  data are applied to the DRaM formula  \Eqn{EnPSolnRot.eq},  
  but errorful    data disrupts the orthonormality of $R_{\opt}$.  Applying the Bar-Itzhack
  or equivalent SVD rotation correction to the deformed rotation produces
  an improved, and orthonormal, rotation matrix that deviates from the
  gold standard minimizing rotation, but is very close, as we see in  Table  \ref{dataTable.tbl}.
  
  The OnP exact-data solution, in hindsight, is just the   the top two lines of the
  EnP solution, or equivalently, what is left when the data for the $``\text{w}''$ 
  components are set to zero.  However, we did actually solve the OnP case
  \emph{before} we looked at the simpler EnP case, and the details of that
  solution are amusing, if not particularly essential, so we add an outline of
  that process here for completeness.  Starting with  the $R(q_{ij})$ inserted
  into the OnP loss  \Eqn{3D2DPoseLSQ.eq},    
  and computing the corresponding
  algebraic  function having only $(\text{ux},\ldots)$ and $(\text{vx},\ldots)$
  components, but no $(\text{wx},\ldots)$ components,\\[-0.15in]
       \begin{equation*} S_{\text{OnP}} (q_{ij}; xx,xy,xz,\ldots;ux,vx,,\ldots)\ , \\[-0.1in]
        \end{equation*}
  the question is what combination of derivatives of  $S_{\text{OnP}} (q_{ij})$
  and the constraints  \Eqn{TheqqConstraints.Eq}  
   is required to produce
  the DRaM solution equal to the top two rows of    \Eqn{EnPSolnRot.eq}, 
  The answer, again following from extensive experimentation with the
  possible candidates, was again to include all 10 derivatives with respect
  to $q_{ij}$, despite the issue of cross-dependencies, and to impose
 the   \emph{first four} of the constraints on the $q_{ij}$ given in 
   \Eqn{TheqqConstraints.Eq},  
 that is, all the constraints containing $q_{00}$,
 and not just the $q_{00} + q_{11} + q_{22} + q_{33} =1$  constraint 
 that sufficed for EnP (any set   containing any single $q_{ii}$ would work as well). 
 The resulting  Mathematica solver code becomes 
 \vspace{-0.1in}
  \begin{equation*} \label{the2DMatchSolver.eq}  
 \begin{aligned}
   & \verb| the3D2DAdjMatchSolns = |  \\
   &\verb|  Module[{eqn = | S_{ \text{OnP}} (q_{ij})\verb|},|  \\
   &\verb|   Solve[ { |\\
   &\verb|     D[eqn, q00] == 0, D[eqn, q11] == 0, |\\
   &\verb|     D[eqn, q22] == 0, D[eqn, q33] == 0,  |\\
   &\verb|     D[eqn, q01] == 0, D[eqn, q02] == 0, |\\
   &\verb|     D[eqn, q03] == 0, D[eqn, q23] == 0,  |\\ 
   &\verb|     D[eqn, q13] == 0, D[eqn, q12] == 0,  |\\
   &\verb|           q00 + q11 + q22 + q33   == 1,   |\\
   &\verb|              q00 q11 == q01 q01,  |\\
   &\verb|              q00 q22 == q02 q02, |\\
   &\verb|              q00 q33 == q03 q03 },   |\\
   &\verb|   {q00, q11, q22, q33, q01, q02, q03, |\\
   &\verb|                        q23, q13, q12}]] | .
  \end{aligned}   
\end{equation*}

\newpage 

  The solver produces a  solution that  is a single list of the 10 $q_{ij}$ rules as ratios of the
  cross-covariances with square-root containing algebraic expressions up to 9MB in length.
  However, substituting the solution values into   \Eqn{qadjFullRot.eq},   
  we find that the first two
  rows, the only ones we \emph{used} in the OnP loss function, simplify  into
  the OnP DRaM solutions corresponding to the top two rows of 
   \Eqn{EnPSolnRot.eq}.  
  The  huge expressions containing enormous square roots have large parts that simply
  cancel out.  In the 3rd row of  \Eqn{qadjFullRot.eq},  
  the enormous parts \emph{add up},
  and that row is nonsense, as we might expect.   However, as  noted in
  \Eqn{OnPPose3rdRowB.eq}, 
  the  (error-free) cross-product of the top two rows converts into an elegant
  DRaM expression for the third row, completing a full  $3 \times 3$
  orthonormal rotation matrix determining the pose from exact OnP data,
  while the usual issues are again present for noisy data.  We observed that the OnP
  solution takes 7 seconds to solve and 33 seconds to assemble the $q_{ij}$ into
  a $2 \times 3$ rotation in the DRaM orthographic form.


 \onecolumn
 

\section{Mathematica Algorithms for the DRaM Method}
 \label{DRaM-algorithm-code.app}

\vspace*{+1.0in}

 \subsection{Quaternion Manipulation Utilities}
 
\vspace*{+0.75in}

      \begin{tabular}{|p{\textwidth}|} 
\hline
\begin{footnotesize}
\begin{verbatim}
  qqDiff[qtest_, qref_] := (2 ArcCos[Min[1.,Abs[qtest.qref]]])/Degree
  
  frobDiff[mat1_, mat2_] := Tr[(mat1-mat2).Transpose[mat1-mat2]]
  
  makeq0plus[q_]:=If[q[[1]]<0,-q,q,q]
  
  pick4DQAdj[adjugate_]:=
      Module[
         {kmax=Position[Abs/@Diagonal[adjugate],Max[Abs/@Diagonal[adjugate]]][[1,1]]},
            Normalize[adjugate[[kmax]]]]

qrotsym = { { q0^2+q1^2-q2^2-q3^2,  2 q1 q2-2 q0 q3,  2 q0 q2+2 q1 q3},
           { 2 q1 q2+2 q0 q3,  q0^2-q1^2+q2^2-q3^2,  -2 q0 q1+2 q2 q3},
           { -2 q0 q2+2 q1 q3,  2 q0 q1+2 q2 q3,  q0^2-q1^2-q2^2+q3^2} }

qqrotsym =  {  q00+q11-q22-q33,  -2 q03 + 2 q12,  2 q02 + 2 q13},
            {  2 q03 + 2 q12,  q00-q11+q22-q33,  -2 q01 + 2 q23},
            {  -2 q02 + 2 q13, 2 q01 + 2 q23,  q00-q11-q22+q33}}
           
QuatToRot[{q0_,q1_,q2_,q3_}] :=
    { {q0^2+q1^2-q2^2-q3^2, 2 q1 q2 - 2 q0 q3, 2 q0 q2 + 2 q1 q3},
      {2 q1 q2 + 2 q0 q3, q0^2-q1^2+q2^2-q3^2, -2 q0 q1+ 2 q2 q3},
      {-2 q0 q2 + 2 q1 q3, 2 q0 q1 + 2 q2 q3, q0^2-q1^2-q2^2+q3^2} }
     
RotToQuat[rot33_] :=   Module[{eig,adjugate,mMat44},
     mMat44=makeMxxmat[Transpose[rot33]];
     eig=Max[Eigenvalues[mMat44]];
     adjugate=Adjugate[mMat44-eig IdentityMatrix[4]];
     makeq0Plus[pick4DQAdj[adjugate]]]
      \end{verbatim}
\end{footnotesize} \\
\hline 
\end{tabular}

\newpage

\subsection{Data Simulation}

\vspace{1.0in}
  \begin{tabular}{|p{\textwidth}|} 
\hline
\begin{footnotesize}
\begin{verbatim}
sigmaSNRFun[tz_,SNR_:70.,w_:2.]:=(Exp[-(SNR/20.)] w)/tz
   (*  For PnP, if tz = 6, SNR 70,  then sigma = 0.01. *)
   
SeedRandom[1357] ;            
RandomReal[{-1, 1}] (*  = -0.75708 (EVALUATE THIS to check Random initialization.) *)

pquat1 = Module[{th = 21.5 Degree, nhat = Normalize[{1.,2.,4.}]},
                Join[{Cos[th/2]},Sin[th/2] nhat]]

prot1 = QuatToRot[pquat1] ;
(* %  Check that initial rotation is reproduceable:
%\begin{array}{ccc}
% 0.933731 & -0.313282 & 0.173208 \\
% 0.326535 & 0.943671 & -0.0534695 \\
% -0.1467 & 0.106485 & 0.983433 \\
%\end{array}
   *)

(*  Settable variables  s = std deviation, camera displacement {tx,ty,tz}  *)

{ pcloud1, protcloud1, protcloudErr1, protTcloud1, 
             portho1, porthoErr1, portho3DErr1  pproj1, pprojErr1} =
             
Module[{npts = 8, rot = prot1, orthoZ=1.0, cloud, cm,              
       rotcloud, protcloudErr, rotTcloud , ortho, orthoErr, ortho3DErr, proj, projErr},
cloud = RandomReal[{-1,1},{npts,3}];
cm = Mean[cloud];
cloud=(#1-cm&)/@cloud;
rotcloud = (rot.#)&/@cloud; 
rotcloudErr  =(s*RandomVariate[NormalDistribution[0,1],{3}]+#)&/@rotcloud;
rotTcloud = Table[(rotcloud[[k]] + {tx,ty,tz}),{k,1,npts}] ;
ortho  = (#[[1;;2]]&)/@rotcloud; 
orthoErr =(s*RandomVariate[NormalDistribution[0,1],{2}]+#)&/@ortho;
ortho3DErr ={#[[1]],#[[2]],orthoZ}&/@orthoErr;
proj  =Table[rotTcloud[[k]][[1;;2]]/rotTcloud[[k]][[3]],{k,1,npts}] ;
projErr = (s*RandomVariate[NormalDistribution[0,1],{2}]+#)&/@proj;
{ cloud, rotcloud, rotcloudErr, rotTcloud, ortho, orthoErr, ortho3DErr, proj, projErr}];
 
 \end{verbatim}
\end{footnotesize}\\ 
\hline 
\end{tabular}

\newpage

\subsection{Profile matrices for quaternion eigensystem.}
      \begin{tabular}{|p{\textwidth}|} 
\hline
\begin{footnotesize}
\begin{verbatim}

(* For the Max eigenvalue quaternion eigensystem characteristic equation. *)
    makeMxxmat[Exxmat_] := 
         Module[{ Exx = Exxmat[[1, 1]], Exy = Exxmat[[1, 2]],  Exz = Exxmat[[1, 3]], 
                  Eyx = Exxmat[[2, 1]], Eyy = Exxmat[[2, 2]],  Eyz = Exxmat[[2, 3]], 
                  Ezx = Exxmat[[3, 1]], Ezy = Exxmat[[3, 2]],  Ezz = Exxmat[[3, 3]]},
    {{Exx + Eyy + Ezz,      Eyz - Ezy,        Ezx - Exz,         Exy - Eyx}, 
    {    Eyz - Ezy,     Exx - Eyy - Ezz,      Exy + Eyx,         Ezx + Exz},
    {    Ezx - Exz,         Exy + Eyx,    -Exx + Eyy - Ezz,      Eyz + Ezy},
    {    Exy - Eyx,         Ezx + Exz,        Eyz + Ezy,     -Exx - Eyy + Ezz}}]
      
(* Bar-Itzhack Method I: partial rotation variation on the characterisic equation. *)
 makeMxx23mat[Exx23mat_]  :=
    Module[{ Exx=Exx23mat[[1,1]], Exy=Exx23mat[[1,2]],  Exz=Exx23mat[[1,3]], 
                   Eyx=Exx23mat[[2,1]], Eyy=Exx23mat[[2,2]],  Eyz=Exx23mat[[2,3]]},
        {{Exx+Eyy,     Eyz,     -Exz,     Exy-Eyx},
         {  Eyz,     Exx-Eyy,   Exy+Eyx,    Exz},
         { -Exz,     Exy+Eyx,  -Exx+Eyy,    Eyz},
         {Exy-Eyx,     Exz,      Eyz,    -Exx-Eyy}}]
    
 (* For the Min eigenvalue quaternion eigensystem characteristic equation. *)
  (* (Atranspose . A)  summed over 'i' *)
buildFullEnPBmat[testVecs_,targetVecs_ ] :=
       Module[{npts = Length[testVecs],ithAMat, 
         aMatFun = Function[{ x,y,z,xx,yy,zz},  
                       {{0,-x +xx ,-y+yy,-z+zz},
                        {x-xx,0,z+zz,-y-yy},
                        {y-yy,-z-zz,0,x+xx}, 
                        {z-zz,y+yy,-x-xx,0}}] },
        Sum[ ithAMat = 
                 aMatFun[ testVecs[[i,1]],testVecs[[i,2]],testVecs[[i,3]],
                          targetVecs[[i,1]],targetVecs[[i,2]],targetVecs[[i,3]]];
                Transpose[ithAMat] . ithAMat,
                  {i,1,npts}]]
 \end{verbatim}
\end{footnotesize} \\
\hline 
\end{tabular}

%


\subsection{The EnP, OnP, and PnP loss functions}


     \begin{tabular}{|p{\textwidth}|} 
\hline
\begin{footnotesize}
\begin{verbatim}


  (* For ArgMin, standard  RMSD Loss  {EnP)  *)
loss3DRMSDFunction [rot_,cloud_, rotcloud_] :=
     Module[{npts= Length[cloud],x,y,z,u,v,w,term },
        1./npts Sum[ {x,y,z} = cloud[[k]];    {u,v,w} = rotcloud[[k]]; 
                               term =   rot . {x,y,z}  - {u,v,w};
                               term . term,
                                  {k,1,npts}]]

(* For ArgMin, standard Ortho Loss  {OnP) *)
loss3DOrthoFunction [rot_,cloud_,ortho_ ] :=
     Module[{npts= Length[cloud], p23rot = rot[[1;;2]], x,y,z,u,v,term},
            1./npts Sum[  x,y,z} = cloud[[k]];       {u,v} = ortho[[k]];
                                   term =  p23rot . {x,y,z}  - {u,v};
                                   term . term,
                                      {k,1,npts}]]

(* For ArgMin, standard z=1 projected image Loss (PnP), given rot and tx,ty,tz  *)
lossDiv3DwTFunction [rot_, t3_, cloud_, img_ ] :=
 Module[{npts = Length[cloud], p23rot = rot[[1 ;; 2]],   denrot = rot[[3]], 
                       x, y, z, u, v, term, txy, tz },
  txy = t3[[1 ;; 2]]; tz = t3[[3]];
  1./npts Sum[{x, y, z} = cloud[[k]]; {u, v} = img[[k]];
                   term =  (p23rot . {x, y, z} + txy)/( denrot . {x, y, z} + tz ) - {u, v};
                   term . term,  
                         {k, 1, npts}]]
\end{verbatim}
\end{footnotesize}\\ 
\hline 
\end{tabular}

 \subsection{The Gold Standards: ArgMin is the Basis of Comparison}   
  \vspace{2in}
  
      \begin{tabular}{|p{\textwidth}|} 
\hline
\begin{footnotesize}
\begin{verbatim}
  
 (* Supplied: the least squares loss functions: and quaternion form R(q) : 
      qrotsym =
      { {q0^2+q1^2-q2^2-q3^2,2 q1 q2-2 q0 q3,2 q0 q2+2 q1 q3},
       {2 q1 q2+2 q0 q3,q0^2-q1^2+q2^2-q3^2,-2 q0 q1+2 q2 q3},
      {-2 q0 q2+2 q1 q3,2 q0 q1+2 q2 q3,q0^2-q1^2-q2^2+q3^2} } *)
  
  (* EnP  and OnP  and PnP  least squares ArgMin  pose estimates *)
  
 getRMSDArgMin[cloud_,rotcloud_ ]:= Module[{ ruleQ, quat, rot },
           ruleQ = FindMinimum[{loss3DRMSDFunction[qrotsym, cloud, rotcloud_],
                           q0^2+q1^2+q2^2+q3^2 ==1}, {{q0,1},{q1,0},{q2,0},{q3,0}},
                           AccuracyGoal -> 12]   [[2]];
           quat = {q0,q1,q2,q3}/.ruleQ;
           rot = QuatToRot[quat];
           {makeq0Plus[quat], rot} ]
           
 getOrthoArgMin[cloud_,ortho_ ] := Module[{ ruleQ, quat, rot },
           ruleQ = FindMinimum[{loss3DOrthoFunction[qrotsym, cloud, ortho ],
                           q0^2+q1^2+q2^2+q3^2 ==1},{{q0,1},{q1,0},{q2,0},{q3,0}},
                            AccuracyGoal -> 12]  [[2]];
           quat = {q0,q1,q2,q3}/.ruleQ;
           rot = QuatToRot[quat];
           {makeq0Plus[quat], rot} ]
    
(* Note: typical  projTS -> 
           pprojErr1 /.{tx->0,ty->0,tz->6.0, s->sigma or n*lhmSigmaFun[tz,70.,w] *)    
             
 getProjArgMin[cloud_,projTS_] :=   Module[{loss,ruleTR,trans,quat,rot},
      {loss, ruleTR }= FindMinimum[
                   {lossDiv3DwTFunction[qrotsym,{tx,ty,tz},cloud,projTS],
                        q0^2+q1^2+q2^2+q3^2 ==1},
                          {{q0, 1}, {q1, 0}, {q2, 0}, {q3, 0}, {tx, 2}, {ty, 3}, {tz, 7}}, 
                            AccuracyGoal -> 12];                        
           trans= {tx,ty,tz}/.ruleTR; 
           quat = {q0,q1,q2,q3}/.ruleTR;
           rot = QuatToRot[quat];
           {makeq0Plus[quat], rot, trans}]
            
   
 \end{verbatim}
\end{footnotesize}\\
\hline 
\end{tabular}

\newpage
 
\subsection{Rotation Correction}

\vspace{2in}

    \begin{tabular}{|p{\textwidth}|} 
\hline
\begin{footnotesize}
\begin{verbatim}
  
  (*  ROTATION EXTENSION:  2 x 3  -> 3 x 3.  Extend top two perfect rows of a
       rotation matrix to 3x3 orthonormal matrix. *)
  
  getRotFrom2x3[P2x3_] := Module[{cross = Normalize[ Cross[P2x3[[1]],P2x3[[2]]] ]},
              Append[P2x3, cross]]

 (* ROTATION REFINEMENT:  Methods that obtain a full orthonormal rotation matrix
       from any  3 x 3 or  2x3 matrix approximating a rotation.
       Bar-Itzhack eigensystem method 1,3 and equivalent square 3D SVD. 
       Bar-Itzhack eigensystem method 2,3 and equivalent rectangular SVD(2x3) 
     NOTE:  The algorithms match  A.Transpose[B],  so if a rotation matrix is
       input, its TRANSPOSE is processed in the algorithm, which means
       these functions are not the same as the corresponding Pose Estimators. *)

getBestQuatRot[Rapprox_] := Module[{Rmat, Mmat, eigopt, quat},
       Rmat = Transpose[Rapprox];
       Mmat = makeMxxmat[Rmat], 
       eigopt = Max[Eigenvalues[Mmat]];
       quat = pick4DQAdj[Adjugate[Mmat - eigopt IdentityMatrix[4]]];
       QuatToRot[quat]]
  
getBestSVDRot[Rapprox] :=  Module[{ Rmat , uu, ss, vv, dd},
     Rmat = Transpose[Rapprox];
    { uu, ss, vv} = SingularValueDecomposition[Rmat];
     dd = DiagonalMatrix[
    Table[If[i == dim, Sign[Det[uu] Det[vv]], 1], {i, 1, dim}]];
     vv . dd . Transpose[uu]]

  (*  Best rotation for 2 x 3  Matrices: Bar-Itzhack Method I  and 2 x 3 singular values;
       Note that both output the TRANSPOSE of the result in a slightly different
       way than the transposed rotation is handled for 3x3 above. *)
  
getHorn23Rot[R23approx_] := 
 Module[{Mmat = makeMxx23mat[R23approx], eigopt, quat},
   eigopt = Max[Eigenvalues[Mmat]];
  quat = pick4DQAdj[Adjugate[Mmat - eigopt IdentityMatrix[4]]];
  Transpose[QuatToRot[quat]]]
  
getSVD23Rot[R23approx_] := Module[{P23=R23approx[[1;;2]],uu,vv,ss,dd,r1,r2,r3,rotOpt},
        {uu,ss,vv}=SingularValueDecomposition[P23];
        dd={{1,0,0},{0,1,0}};
        {r1,r2}=uu.dd.Transpose[vv];   (* Transpose of standard SVD output *)
        r3=Normalize[ Cross[r1, r2];
        rotOpt={r1,r2,r3}]
 \end{verbatim}
\end{footnotesize}\\
\hline 
\end{tabular}

\newpage

  \subsection{The List of  RMSD-class EnP Pose Discovery Functions}
  
   \vspace{ 0.75in}
  
  These   algorithms  produce agreement with
  the \ArgMin Gold Standard for EnP data \emph{with or without noise}.
  
 \vspace{0.75in}
  
     \begin{tabular}{|p{\textwidth}|} 
\hline
\begin{footnotesize}
\begin{verbatim}
getRMSDArgMin[cloud_, rotcloud_ ] :=  Module[{loss,ruleQ, quat, rot}, 
  {loss, ruleQ} = FindMinimum[{loss3DRMSDFunction[qrotsym, cloud, rotcloud], 
         q0^2 + q1^2 + q2^2 + q3^2 == 1}, {{q0, 1}, {q1, 0}, {q2, 0}, {q3, 0}},
        AccuracyGoal -> 12] ;
        quat = {q0, q1, q2, q3} /. ruleQ; 
       {makeq0Plus[quat],  QuatToRot[quat]}]
  
getHebertRotXY[cloud_, target_] := Module[{ Bmat, eigopt, quat},
       Bmat = buildFullEnPBmat[cloud, target];
      eigopt = Chop[ Min[Eigenvalues[Bmat]]];
      quat = pick4DQAdj[Adjugate[Bmat - eigopt IdentityMatrix[4]]];
      QuatToRot[quat]]
  
getHornRotXY[ cloud_, target_] :=  Module[{eigopt, quat, adjugate, eMat33, mMat44},
       eMat33 = Transpose[cloud] . target;
       mMat44 = makeMxxmat[eMat33];
       eigopt = Max[Eigenvalues[mMat44]];
       adjugate = Adjugate[mMat44 - eigopt*IdentityMatrix[4]]; 
        (* The pick4DQAdj[ ]  normalizes *)
        quat = makeq0Plus[ pick4DQAdj[adjugate] ] ;
        QuatToRot[quat]]
   
getSVDRotXY[cloud_, target_] := Module[{uu, ss, vv, dd, eMat, dim},
        eMat = Transpose[cloud] . target;  dim = Dimensions[eMat][[1]];
        {uu, ss, vv} = SingularValueDecomposition[eMat];
       dd = DiagonalMatrix[Table[If[i == dim, Sign[Det[uu] Det[vv]], 1], {i, 1, dim}]];
       vv . dd . Transpose[uu]]
  
 (* HNN has two options: "Plus root" and "Minus root":
     This First version needs inverse of  eMat, which is a problem if Det[eMat]=0 *)
getHHNRotPXY[cloud_, target_] := Module[{eMat, rootEtE, rotInv, rotOpt},
         eMat = Transpose[cloud] . target ;
         rotInv = Inverse[eMat];
          rootEtE = MatrixPower[Transpose[eMat] . eMat, +1/2];
          rotOpt = rootEtE . rotInv] 
           
(* Our Preferred version does not need inverse of eMat, just inverse of  
        the S factor and a Transpose. *)
getHHNRotMXY[cloud_, target_] := Module[{invRootEtE, rotOpt, eMat},
         eMat = Transpose[cloud] . target ; 
         invRootEtE = MatrixPower[Transpose[eMat] . eMat, -1/2];
         rotOpt = invRootEtE . Transpose[eMat]]       
\end{verbatim} \vspace{0.1in}
\end{footnotesize}\\ 
\hline 
\end{tabular}
    
    \newpage
    
    \subsection{Alternative forms with single cross-covariance matrix argument.}
    
    \vspace{1.0in}
    
   \noindent
  \begin{tabular}{|p{\textwidth}|} 
\hline
\begin{footnotesize}
\begin{verbatim}
 

(* Native form of quaternion eigenvalue system, corresponding to   
    getHornRotXY[cloud, target]   above. *)
    
getHornRotEmat [Emat_] :=   Module[{Mmat, eigopt, quat}, 
       Mmat = makeMxxmat[Emat];
       eigopt = Max[Eigenvalues[Mmat]]; 
       quat = pick4DQAdj[Adjugate[Mmat - eigopt IdentityMatrix[4]]]; 
       QuatToRot[quat]]
  
(*  Native form of singular value decomposition system, corresponding to  
       getSVDRotXY[cloud, target] above. *)
       
getSVDRotEmat[eMat_] := Module[{dim =Dimensions[eMat][[1]], uu,ss,vv,dd},
          { uu,ss,vv}=SingularValueDecomposition[eMat];
           dd = DiagonalMatrix[Table[If[i==dim,Sign[Det[uu]Det[vv]],1],{i,1,dim}]];
           vv.dd.Transpose[uu]]
            
(*  Native form of HNN matrix algorithm, corresponding to
     getHHNRotPXY[cloud_ target] and  getHHNRotMXY[cloud, target]  above.
 
 (* HNN  Emat  has two options: Plus 1/2 root and Minus 1/2 root *)
       
(*  HNN "Plus  root":This First version needs inverse of eMat, 
which is a problem if Det[eMat]=0 *)

getHHNRotPEmat [eMat_] :=  Module[{rootEtE, rotInv, rotOpt}, 
       rotInv = Inverse[eMat];
       rootEtE = MatrixPower[Transpose[eMat] . eMat, +1/2];
       rotOpt = rootEtE . rotInv]

(*  HNN "Minus root" Our Preferred version does not need inverse of eMat,  
     just inverse of the S factor and a Transpose.*)
     
getHHNRotMEmat [eMat_] :=  Module[{invRootEtE, rotOpt}, 
       invRootEtE = MatrixPower[Transpose[eMat] . eMat, -1/2];
       rotOpt = invRootEtE . Transpose[eMat]]
   \end{verbatim}        
\end{footnotesize}\\
\hline 
\end{tabular}

\newpage
 
\subsection{The List of DRaM class EnP Pose Discovery Functions}
  
  \vspace{ 0.75in}
  
  These functions produce perfect rotation results \emph{only} for perfect,
  error-free inputs, and become non-orthonormal in the presence of error.
  Rotation Correction restores orthonormality, and their accuracy in the
  presence of error is then within a few degrees of the \ArgMin, but never exact.

\vspace{ 0.75in}

%
 \noindent
     \begin{tabular}{|p{\textwidth}|} 
\hline
\begin{footnotesize}
\begin{verbatim}
 getAnyQRDecRot[cloud_, target_] :=  Module[{theS, theT}, 
           {theS, theT} = QRDecomposition[cloud];
           Transpose[target] . Transpose[theS] . Inverse[Transpose[theT]]]
    
(* Reference:  Moore-Penrose PseudoInverse has this implementation *)
myPseudoInverse[matrix_] :=  Module[{MtMForm = Transpose[matrix] . matrix},
             Inverse[MtMForm] . Transpose[matrix]]
             
(* This  extracts the pose matrix using only the  cloud data matrix: *)            
 getAnyPseudoInverseRot[cloud_, target_] :=  Transpose[PseudoInverse[cloud] . target]
    
 getEnPDRaM[cloud_, target_] :=
  Module[{the3DMatchDraM, d0, d11, d12, d13, d21, d22, d23, d31, d32, d33},
   Module[{xx, yy, zz, xy, xz, yz, ux, uy, uz, vx, vy, vz, wx, wy, wz},
    Module[{ xList = Transpose[cloud][[1]], yList = Transpose[cloud][[2]], 
             zList = Transpose[cloud][[3]],
     uList = Transpose[target][[1]],  vList = Transpose[target][[2]], 
            wList = Transpose[target][[3]]}, 
     {xx, yy, zz, xy, xz, yz, ux,  uy, uz, vx, vy, vz , wx, wy, wz} = 
         {xList . xList, yList . yList, zList . zList, 
               xList . yList,  xList . zList, yList . zList,
      uList . xList, uList . yList, uList . zList, 
          vList . xList, vList . yList, vList . zList,
              wList . xList, wList . yList, wList . zList};
     d0 = Det[{{xx, xy, xz}, {xy, yy, yz}, {xz, yz, zz}}];
    d11 = Det[{{xy, yy, yz}, {xz, yz, zz}, {ux, uy, uz}}]; 
    d12 = Det[{{xz, yz, zz}, {xx, xy, xz}, {ux, uy, uz}}]; 
    d13 = Det[{{xx, xy, xz}, {xy, yy, yz}, {ux, uy, uz}}]; 
    d21 = Det[{{xy, yy, yz}, {xz, yz, zz}, {vx, vy, vz}}]; 
    d22 = Det[{{xz, yz, zz}, {xx, xy, xz}, {vx, vy, vz}}]; 
    d23 = Det[{{xx, xy, xz}, {xy, yy, yz}, {vx, vy, vz}}]; 
    d31 = Det[{{xy, yy, yz}, {xz, yz, zz}, {wx, wy, wz}}];
    d32 = Det[{{xz, yz, zz}, {xx, xy, xz}, {wx, wy, wz}}];
    d33 = Det[{{xx, xy, xz}, {xy, yy, yz}, {wx, wy, wz}}];
    the3DMatchDraM = {{d11/d0, d12/d0, d13/d0}, 
                      {d21/d0, d22/d0, d23/d0},
                      {d31/d0, d32/d0, d33/d0}}] ]]
         \end{verbatim}
\end{footnotesize}\\
\hline 
\end{tabular}
 
 \newpage
 
 \vspace{ 0.75in}
 
 Rotation Correction forms for the non-orthonormal DRaM class candidates for the
 EnP pose with  errorful data.   Our convention is to require the transpose when 
 using \verb|getSVDRot[mat]|    or equivalent for rotation correction.
  
 \vspace{ 0.75in}
  
\noindent
     \begin{tabular}{|p{\textwidth}|} 
\hline
\begin{footnotesize}
\begin{verbatim}

(* Using this form, which solves any EnP problem with the cross-covariance
      matrix as the argument, must be called with the transpose rotation
      matrix candidate for Rotation Correction. *)
      
getSVDRot[matrix_] :=  Module[{dim = Dimensions[matrix][[1]], uu, ss, vv, dd},
         {uu, ss, vv} =  SingularValueDecomposition[matrix]; 
         dd = DiagonalMatrix[
                  Table[If[i == dim, Sign[Det[uu] Det[vv]], 1], {i, 1, dim}]]; 
         vv . dd . Transpose[uu]]


(*  Corrected  DRaM changes angular error from 1.78 deg to 1.42 deg  *)
getEnPDRaMRC[cloud_, target_] :=
       getSVDRot[Transpose[getEnPDRaM[ cloud, target]]  ]
       
(*  Corrected  QR Decomposition  changes angular error from 1.78 deg to 1.42 deg  *)       
getAnyQRDecRotRC[cloud_, target_] :=
     getSVDRot[Transpose[getAnyQRDecRot[cloud, target]]]
 
 (*  Corrected  PseudoInverse Map  changes angular error from 1.78 deg to 1.42 deg  *)
getAnyPseudoInverseRotRC[cloud_, target_] :=
    getSVDRot[Transpose[getAnyPseudoInverseRot[cloud, target]]]
 
  \end{verbatim}        
\end{footnotesize}\\
\hline 
\end{tabular}

\clearpage

\subsection{The List of RMSD-class OnP Pose Discovery Functions}

   \vspace{0.75in}

  These are adapted from the EnP class to handle a planar 2D target,
  which always fails to give a reasonable pose estimate.
     
     \vspace{0.75in}
     
     \begin{tabular}{|p{\textwidth}|} 
\hline
\begin{footnotesize}
\begin{verbatim}
getOrthoArgMin[cloud_, ortho_ ] :=   Module[{loss,ruleQ, quat, rot}, 
   {loss, ruleQ } = FindMinimum[{loss3DOrthoFunction[qrotsym,  cloud,  ortho], 
         q0^2 + q1^2 + q2^2 + q3^2 == 1}, {{q0, 1}, {q1, 0}, {q2,  0}, {q3, 0}},
          AccuracyGoal -> 12 ]; 
   quat = {q0, q1, q2, q3} /. ruleQ; 
  {makeq0Plus[quat], QuatToRot[quat]}]
    
getHebert23Rot[cloud_, ortho_] :=  Module[{Bmat, eigopt, quat}, 
     Bmat = buildFullEnPBmat[cloud, ({#1[[1]], #1[[2]], 0} &) /@ ortho]; 
     eigopt = Chop[Min[Eigenvalues[Bmat]]]; 
     quat = pick4DQAdj[Adjugate[Bmat - eigopt IdentityMatrix[4]]]; 
     QuatToRot[quat]]
  
(* This has an obvious alternate  getHorn23Rot[Emat23] form *)
getHorn23RotXY[cloud_, ortho_] :=  Module[{Mmat, Emat23, eigopt, quat},
      Emat23 = Transpose[  Transpose[cloud] . ortho]; 
     Mmat = makeMxx23mat[Emat23];
    eigopt = Max[Eigenvalues[Mmat]]; 
     quat = pick4DQAdj[Adjugate[Mmat - eigopt IdentityMatrix[4]]]; 
    Transpose[QuatToRot[quat]] ]
    
(* This has an obvious alternate  getSVD23Rot[Emat23] form *)    
 getSVD23RotXY[cloud_, ortho_] := 
 Module[{Emat23 = Transpose[Transpose[cloud] . ortho], uu, vv, ss, dd, 
   r1, r2, r3, rotOpt}, {uu, ss, vv} = SingularValueDecomposition[Emat23];
  dd = {{1, 0, 0}, {0, 1, 0}}; {r1, r2} = uu . dd . Transpose[vv];
  r3 = Normalize[Cross[r1, r2]]; rotOpt = {r1, r2, r3}]
   
    (* This  HHN  2x3  version  exploits the PseudoInverse of the 
                 singular matrix  EtE and that works..  *)
 (* This has an obvious alternate  getHNN23Rot[Emat23] form *)           
  getHHN23RotXY [cloud_, ortho_] :=  
  Module[{Emat23, EtE, pseudoinv, rootEtEinv, p23Opt, rot3, rotOpt}, 
      Emat23 = Transpose[Transpose[cloud] . ortho];
      EtE = Transpose[Emat23] . Emat23; 
      pseudoinv = PseudoInverse[EtE]; 
      rootEtEinv = MatrixPower[pseudoinv, +1/2]; 
      p23Opt = Emat23 . rootEtEinv; 
      rot3 = Normalize[Cross[p23Opt[[1]], p23Opt[[2]]]]; 
      rotOpt = {p23Opt[[1]], p23Opt[[2]], rot3}]     
 \end{verbatim}
\end{footnotesize}\\
\hline 
\end{tabular}

  \newpage
  
\subsection{The List of DRaM class OnP Pose Discovery Functions}

 \vspace{0.75in}

  These functions produce perfect rotation results \emph{only} for perfect,
  error-free inputs, and become non-orthonormal in the presence of error.
  Rotation Correction restores orthonormality, and their accuracy in the
  presence of error is then within a few degrees of the \ArgMin, but never exact.
  
   \vspace{0.75in}


     \begin{tabular}{|p{\textwidth}|} 
\hline
\begin{footnotesize}
\begin{verbatim}
(* These are universal, for 3x3 targets or 2x3 ortho data targets.  *)
 getAnyQRDecRot[cloud_, ortho_] :=  Module[{theS, theT}, 
           {theS, theT} = QRDecomposition[cloud];
           Transpose[ortho] . Transpose[theS] . Inverse[Transpose[theT]]]
       
(* These are universal, for 3x3 targets or 2x3 ortho data targets.  *)
 getAnyPseudoInverseRot[cloud_, ortho_] :=  Transpose[PseudoInverse[cloud] . ortho]

   
 getOnPDRaM[cloud_, ortho_] :=
 Module[{the3DposeInit, d0, d11, d12, d13, d21, d22, d23, dd31, dd32, dd33},
  Module[{xx, yy, zz, xy, xz, yz, ux, uy, uz, vx, vy, vz},
   Module[{
      xList = Transpose[cloud][[1]], 
            yList = Transpose[cloud][[2]], 
                 zList = Transpose[cloud][[3]],
      uList = Transpose[ortho][[1]],   
            vList = Transpose[ortho][[2]]}, 
      {xx, yy, zz,   xy, xz, yz,   ux, uy, uz,    vx, vy, vz } = 
          {xList . xList, yList . yList, zList . zList,  
             xList . yList, xList . zList, yList . zList,
               uList . xList, uList . yList, uList . zList, 
                   vList . xList, vList . yList, vList . zList};
    d0 = Det[{{xx, xy, xz}, {xy, yy, yz}, {xz, yz, zz}}];
    d11 = Det[{{xy, yy, yz}, {xz, yz, zz}, {ux, uy, uz}}]; 
    d12 = Det[{{xz, yz, zz}, {xx, xy, xz}, {ux, uy, uz}}]; 
    d13 = Det[{{xx, xy, xz}, {xy, yy, yz}, {ux, uy, uz}}]; 
    d21 = Det[{{xy, yy, yz}, {xz, yz, zz}, {vx, vy, vz}}]; 
    d22 = Det[{{xz, yz, zz}, {xx, xy, xz}, {vx, vy, vz}}]; 
    d23 = Det[{{xx, xy, xz}, {xy, yy, yz}, {vx, vy, vz}}]; 
    dd31 = Det[{{xx, xy, xz}, {ux, uy, uz}, {vx, vy, vz}}];
    dd32 = Det[{{xy, yy, yz}, {ux, uy, uz}, {vx, vy, vz}}];
    dd33 = Det[{{xz, yz, zz}, {ux, uy, uz}, {vx, vy, vz}}];
    the3DposeInit  =
                    {{d11/d0,  d12/d0,  d13/d0}, 
                     {d21/d0,  d22/d0,  d23/d0}, 
                     {dd31/d0, dd32/d0, dd33/d0}}] ]] 
                  \end{verbatim}
\end{footnotesize}\\
\hline 
\end{tabular}

\clearpage

 \vspace{0.75in}

 Rotation Correction forms for the non-orthonormal DRaM class candidates for the
 OnP pose with  errorful data.   Our convention is to require the transpose when 
 using \verb|getSVDRot[mat]|    or equivalent for rotation correction.
  
  \vspace{0.75in}

\noindent
     \begin{tabular}{|p{\textwidth}|} 
\hline
\begin{footnotesize}
\begin{verbatim}

(* This form, which solves any EnP problem with the cross-covariance
      matrix as the argument, must be called with the transpose of
      a 3x3 rotation  matrix candidate for Rotation Correction. *)
      
getSVDRot[matrix_] :=  Module[{dim = Dimensions[matrix][[1]], uu, ss, vv, dd},
         {uu, ss, vv} =  SingularValueDecomposition[matrix]; 
         dd = DiagonalMatrix[
                  Table[If[i == dim, Sign[Det[uu] Det[vv]], 1], {i, 1, dim}]]; 
         vv . dd . Transpose[uu]]

(* This form takes a 2x3 partial rotation matrix and correct it, no transpose.
    Note the inversion of the final SVD matrix order. *)
getSVD23Rot[Emat23_] :=  Module[{uu, vv, ss, dd, r1, r2, r3, rotOpt}, 
     {uu, ss, vv} =   SingularValueDecomposition[Emat23]; 
    dd = {{1, 0, 0}, {0, 1, 0}}; {r1, r2} = uu . dd . Transpose[vv]; 
     r3 = Normalize[Cross[ r1, r2]; 
     rotOpt = {r1, r2, r3}]
 
(*  Corrected  DRaM changes angular error from nonsense  to 2.85 deg  *)
getOnPDRaMRC[cloud_, ortho] :=
       getSVDRot[Transpose[getOnPDRaM[ cloud, ortho]]  ]
       
(*  Corrected  QR Decomposition  changes angular error from  nonsense  to 2.85 deg  *) 
getOnPQRDecRotRC[cloud_, ortho] :=
     getSVD23Rot[getAnyQRDecRot[cloud, ortho]]
 
 (*  Corrected  PseudoInverse Map  changes angular error from  nonsense  to 2.85 deg  *)
getOnPPseudoInverseRotRC[cloud_, ortho_] :=
       getSVD23Rot[getAnyPseudoInverseRot[cloud, ortho]]
 
  \end{verbatim}        
\end{footnotesize}\\
\hline 
\end{tabular}

\newpage

\subsection{The 3D:3D  EnP DRaM Solution}

 \vspace{0.75in}

     \begin{tabular}{|p{\textwidth}|} 
\hline
\begin{footnotesize}
\begin{verbatim}
poseLSQEnP = 
 Module[{ rotq = {{q00 + q11 - q22 - q33, -2 q03 + 2 q12,  2 q02 + 2 q13},
                            {2 q03 + 2 q12,  q00 - q11 + q22 - q33, -2 q01 + 2 q23}, 
                            {-2 q02 + 2 q13, 2 q01 + 2 q23, q00 - q11 - q22 + q33}}},
   Expand[(rotq . {x, y, z} - {u, v, w}) .(*dot*)(rotq . {x, y, z} - {u, v, w})  ] /. 
                {x^2 -> xx,  x y -> xy, x z -> xz, y x -> xy, y^2 -> yy, y z -> yz, 
                z x -> xz,  z y -> yz, z^2 -> zz,
                 u x -> ux, u y -> uy, u z -> uz, v x -> vx, v y -> vy,  v z -> vz,  
                 x w -> wx,  y w -> wy,  z w -> wz,  u^2 -> uu, v^2 -> vv}] // 
                 Collect[#, {q00, q11, q22, q33, q01, q02, q03, q23, q13, q12}] &
 
 Out[] :=   uu+vv+q12 (-4 uy-4 vx)+w^2+q23 (-4 vz-4 wy+8 q13 xy+8 q12 xz)+
 q03^2 (4 xx+4 yy)+q12^2 (4 xx+4 yy)+q33 (2 ux+2 vy-2 wz-8 q12 xy-8 q02 xz+
 8 q01 yz)+q13 (-4 uz-4 wx+8 q12 yz)+q03 (4 uy-4 vx+8 q23 xz+q12 (8 xx-8 yy)-
 8 q13 yz)+q01 (4 vz-4 wy-8 q02 xy+8 q13 xy-8 q03 xz-8 q12 xz+q23 (8 yy-8 zz))+
 q22 (2 ux-2 vy+2 wz+8 q03 xy-8 q13 xz-8 q01 yz+q33 (2 xx-2 yy-2 zz))+
 q00^2 (xx+yy+zz)+q11^2 (xx+yy+zz)+q22^2 (xx+yy+zz)+q33^2 (xx+yy+zz)+
 q02^2 (4 xx+4 zz)+q13^2 (4 xx+4 zz)+q01^2 (4 yy+4 zz)+q23^2 (4 yy+4 zz)+
 q11 (-2 ux+2 vy+2 wz-8 q03 xy+8 q02 xz-8 q23 yz+q33 (-2 xx+2 yy-2 zz)+
 q22 (-2 xx-2 yy+2 zz))+q00 (-2 ux-2 vy-2 wz+8 q12 xy+8 q13 xz+8 q23 yz+
 q11 (2 xx-2 yy-2 zz)+q22 (-2 xx+2 yy-2 zz)+q33 (-2 xx-2 yy+2 zz))+
 q02 (-4 uz+4 wx-8 q23 xy-8 q03 yz+8 q12 yz+q13 (-8 xx+8 zz)) 
 
  (*   ~ 0.5  seconds to Solve,   ~ 9 seconds to Simplify *)
  theAdjSolnsEnP = 
     Module[{eqn = poseLSQEnP}, 
         Solve[{D[eqn, q00] == 0, D[eqn, q11] == 0, D[eqn, q22] == 0, 
                    D[eqn, q33] == 0, D[eqn, q01] == 0, D[eqn, q02] == 0, 
                    D[eqn, q03] == 0, D[eqn, q23] == 0, D[eqn, q13] == 0, 
                    D[eqn, q12] == 0, q00 + q11 + q22 + q33 == 1}, 
                    {q00, q11, q22,  q33, q01, q02, q03, q23, q13, q12}]][[1]] /Simplify ]
  
  qqrotsym /. theAdjSolnsEnP // Simplify
  
  Out[]:=  (* The 3 x 3 EnP DRaM rotation matrix *)
  
                  \end{verbatim}
\end{footnotesize}\\
\hline 
\end{tabular}
  
  \newpage

\subsection{The 3D:2D   Orthographic OnP DRaM Solution}

 \vspace{0.75in}

     \begin{tabular}{|p{\textwidth}|} 
\hline
\begin{footnotesize}
\begin{verbatim}

 poseLSQOnP=Expand[({{q00+q11-q22-q33,-2 q03+2 q12,2 q02+2 q13},
            {2 q03+2 q12,q00-q11+q22-q33,-2 q01+2 q23}}.{x,y,z}-{u,v}).(*dot*)
    ({{q00+q11-q22-q33,-2 q03+2 q12,2 q02+2 q13},
        {2 q03+2 q12,q00-q11+q22-q33,-2 q01+2 q23}} . {x,y,z} - {u,v})]/.
           {x^2->xx,x y->xy,x z->xz,y x->xy,y^2->yy,y z->yz,z x->xz,z y->yz,z^2->zz,
             u x->ux,u y->uy,u z->uz,v x->vx,v y->vy,v z->vz,u^2->uu,v^2->vv}//
       Collect[#,{q00,q11,q22,q33,q01,q02,q03,q23,q13,q12}]&
       
Out[]:= uu+vv+q12 (-4 uy-4 vx)+q23 (-4 vz+8 q12 xz)+q00^2 (xx+yy)+q11^2 (xx+yy)+
q22^2 (xx+yy)+q33^2 (xx+yy)+q03^2 (4 xx+4 yy)+q12^2 (4 xx+4 yy)+q13 (-4 uz+
8 q12 yz)+q03 (4 uy-4 vx+8 q23 xz+q12 (8 xx-8 yy)-8 q13 yz)+q33 (2 ux+
2 vy-8 q12 xy-4 q02 xz-4 q13 xz+4 q01 yz-4 q23 yz)+q11 (-2 ux+2 vy-8 q03 xy+
4 q02 xz+4 q13 xz+q22 (-2 xx-2 yy)+q33 (-2 xx+2 yy)+4 q01 yz-4 q23 yz)+
q22 (2 ux-2 vy+8 q03 xy-4 q02 xz-4 q13 xz+q33 (2 xx-2 yy)-4 q01 yz+4 q23 yz)+
q00 (-2 ux-2 vy+8 q12 xy+4 q02 xz+4 q13 xz+q33 (-2 xx-2 yy)+q11 (2 xx-2 yy)+
q22 (-2 xx+2 yy)-4 q01 yz+4 q23 yz)+4 q01^2 zz+4 q02^2 zz+4 q13^2 zz+
4 q23^2 zz+q02 (-4 uz-8 q03 yz+8 q12 yz+8 q13 zz)+
q01 (4 vz-8 q03 xz-8 q12 xz-8 q23 zz)
                                 
 (* ~ 7 seconds to Solve,  ~ 33 seconds to Simplify *)                                
 the3D2DAdjSolns = 
   Module[{eqn = poseLSQOnP}, 
     Solve[{D[eqn, q00] == 0, D[eqn, q11] == 0, D[eqn, q22] == 0, 
       D[eqn, q33] == 0, D[eqn, q01] == 0, D[eqn, q02] == 0, 
       D[eqn, q03] == 0, D[eqn, q23] == 0, D[eqn, q13] == 0, 
       D[eqn, q12] == 0, q00 + q11 + q22 + q33 == 1, 
       q00 q11 == q01 q01, q00 q22 == q02 q02, q00 q33 == q03 q03},
      {q00, q11, q22, q33, q01, q02, q03, q23, q13, q12}]][[1]]]
      
(* Top row of OnP DRaM: *)
   {q00 + q11 - q22 - q33,  -2 q03 + 2 q12, 2 q02 + 2 q13} /. the3D2DAdjSolns // Simplify 
   
(* Second row of OnP DRaM: *)
   {2 q03 + 2 q12,  q00 - q11 + q22 - q33, -2 q01 + 2 q23} /.  the3D2DAdjSolns  // Simplify 
  
  Out[]:= (* The 2 x 3 OnP DRaM rotation matrix *)

                    \end{verbatim}
                    
\end{footnotesize} \\[-0.0in] \hline 
\end{tabular}

  \newpage
  

\subsection{The rij - based 3D:3D  EnP DRaM Solution}
     \begin{tabular}{|p{\textwidth}|} 
\hline
\begin{footnotesize}
\begin{verbatim}
  (* This least squares form works for EnP and also OnP if we drop 3rd line of rij *)
      
poseLSQEnPRR = 
 Module[{ rot  = {{r11, r12, r13}, {r21, r22, r23}, {r31, r32, r33}}},
   Expand[(rot . {x, y, z} - {u, v, w}) .(*dot*)(rot . {x, y, z} - {u, v, w})] /. 
    {x^2 -> xx, x y -> xy, x z -> xz, y x -> xy, y^2 -> yy,  y z -> yz, z x -> xz, 
    z y -> yz, z^2 -> zz,  u x -> ux, u y -> uy, u z -> uz, v x -> vx, v y -> vy,
    v z -> vz,  x w -> wx,  y w -> wy,  z w -> wz, u^2 -> uu, v^2 -> vv, w^2 -> ww}] // 
  Collect[#, {r11, r12, r13, r21, r22, r23, r31, r32, r33}] & 

Out -> uu - 2 r13 uz + vv - 2 r23 vz + ww - 2 r33 wz + 
  r11^2 xx + r21^2 xx +  r31^2 xx + r11 (-2 ux + 2 r12 xy + 2 r13 xz) + 
  r21 (-2 vx + 2 r22 xy + 2 r23 xz) +  r31 (-2 wx + 2 r32 xy + 2 r33 xz) +
  r12^2 yy + r22^2 yy + r32^2 yy +  r12 (-2 uy + 2 r13 yz) + r22 (-2 vy + 2 r23 yz) +
  r32 (-2 wy + 2 r33 yz) +  r13^2 zz + r23^2 zz + r33^2 zz
  
(*  The solution of the LSQ symbolic form (for perfect data). *)
Timing[
  rijOfxyzuvwRule  = Module[{eqn = poseLSQEnPRR}, 
    Solve[{  D[eqn, r11] == 0, D[eqn, r12] == 0, D[eqn, r13] == 0,
             D[eqn, r21] == 0, D[eqn, r22] == 0, D[eqn, r23] == 0,
             D[eqn, r31] == 0, D[eqn, r32] == 0, D[eqn, r33] == 0},
        {r11, r12, r13, r21, r22, r23, r31, r32, r33}] ] [[1]] //  Simplify ]
          
 Out =->  0.011 sec.
 {r11 -> (uz xz yy - uz xy yz - uy xz yz + ux yz^2 + uy xy zz - ux yy zz)/
                (xz^2 yy - 2 xy xz yz + xx yz^2 + xy^2 zz - xx yy zz), 
  r12 -> (-uz xy xz + uy xz^2 + uz xx yz - ux xz yz - uy xx zz +  ux xy zz)/
                (xz^2 yy - 2 xy xz yz + xx yz^2 + xy^2 zz - xx yy zz), 
  r13 -> (uz xy^2 - uy xy xz - uz xx yy + ux xz yy + uy xx yz - ux xy yz)/
                (xz^2 yy - 2 xy xz yz + xx yz^2 + xy^2 zz - xx yy zz), 
  r21 -> (vz xz yy - vz xy yz - vy xz yz + vx yz^2 + vy xy zz - vx yy zz)/
                 (xz^2 yy - 2 xy xz yz + xx yz^2 + xy^2 zz - xx yy zz), 
  r22 -> (-vz xy xz + vy xz^2 + vz xx yz - vx xz yz - vy xx zz + vx xy zz)/
                 (xz^2 yy - 2 xy xz yz + xx yz^2 + xy^2 zz - xx yy zz), 
  r23 -> (vz xy^2 - vy xy xz - vz xx yy + vx xz yy + vy xx yz - vx xy yz)/
                  (xz^2 yy - 2 xy xz yz + xx yz^2 + xy^2 zz - xx yy zz),
  r31 -> (wz xz yy - wz xy yz - wy xz yz + wx yz^2 + wy xy zz - wx yy zz)/
                   (xz^2 yy - 2 xy xz yz + xx yz^2 + xy^2 zz - xx yy zz), 
  r32 -> (-wz xy xz + wy xz^2 + wz xx yz - wx xz yz - wy xx zz + wx xy zz)/
                   (xz^2 yy - 2 xy xz yz + xx yz^2 + xy^2 zz - xx yy zz), 
  r33 -> (wz xy^2 - wy xy xz - wz xx yy + wx xz yy + wy xx yz - wx xy yz)/
                   (xz^2 yy - 2 xy xz yz + xx yz^2 + xy^2 zz - xx yy zz) }           
               \end{verbatim}
\end{footnotesize}\\
\hline 
\end{tabular}


\subsection{The rij - based 3D:2D  Orthographic OnP DRaM Solution}
     \begin{tabular}{|p{\textwidth}|} 
\hline
\begin{footnotesize}
\begin{verbatim}
 
 (* Same eqn as EnP can be used, omit r3k bottom line. *)
 Timing[
   rijOnPxyzuvRule  = Module[{eqn = poseLSQEnPRR}, 
    Solve[{  D[eqn, r11] == 0,   D[eqn, r12] == 0,   D[eqn, r13] == 0,
                 D[eqn, r21] == 0,   D[eqn, r22] == 0,    D[eqn, r23] == 0 
                    (*D[eqn,r31]==0, D[eqn,r32]==0,  D[eqn,r33]==0*)},
                 {r11, r12, r13, r21, r22, r23}] ] [[1]] //   Simplify ]
 
Out ->  0.0082  sec
{r11 -> (uz xz yy - uz xy yz - uy xz yz + ux yz^2 + uy xy zz - ux yy zz)/
               (xz^2 yy - 2 xy xz yz + xx yz^2 + xy^2 zz - xx yy zz), 
  r12 -> (-uz xy xz + uy xz^2 + uz xx yz - ux xz yz - uy xx zz +  ux xy zz)/
               (xz^2 yy - 2 xy xz yz + xx yz^2 + xy^2 zz - xx yy zz), 
  r13 -> (uz xy^2 - uy xy xz - uz xx yy + ux xz yy + uy xx yz - ux xy yz)/
               (xz^2 yy - 2 xy xz yz + xx yz^2 + xy^2 zz - xx yy zz), 
  r21 -> (vz xz yy - vz xy yz - vy xz yz + vx yz^2 + vy xy zz - vx yy zz)/
               (xz^2 yy - 2 xy xz yz + xx yz^2 + xy^2 zz - xx yy zz), 
  r22 -> (-vz xy xz + vy xz^2 + vz xx yz - vx xz yz - vy xx zz + vx xy zz)/
                (xz^2 yy - 2 xy xz yz + xx yz^2 + xy^2 zz - xx yy zz), 
  r23 -> (vz xy^2 - vy xy xz - vz xx yy + vx xz yy + vy xx yz - vx xy yz)/
                (xz^2 yy - 2 xy xz yz + xx yz^2 + xy^2 zz - xx yy zz)}  
                      \end{verbatim}
\end{footnotesize}\\
\hline 
\end{tabular}

 \comment{     

If Room - replaced by  getHornRotXY  and getSVDRotXY[cloud, target]

getHornRotEmat [Emat_] :=   Module[{Mmat, eigopt, quat}, 
       Mmat = makeMxxmat[Emat];
       eigopt = Max[Eigenvalues[Mmat]]; 
       quat = pick4DQAdj[Adjugate[Mmat - eigopt IdentityMatrix[4]]]; 
      QuatToRot[quat]]

getSVDRotEmat[eMat_] := Module[{dim =Dimensions[eMat][[1]], uu,ss,vv,dd},
          { uu,ss,vv}=SingularValueDecomposition[eMat];
           dd = DiagonalMatrix[Table[If[i==dim,Sign[Det[uu]Det[vv]],1],{i,1,dim}]];
          vv.dd.Transpose[uu]]

=================
 If room - replaced by getHHNRotPXY[cloud_, target_]
                                      getHHNRotMXY[cloud_, target_]
 
 (* HNN  Emat  has two options: Plus 1/2 root and Minus 1/2 root *)

(* 5 HNN "Plus  root":This First version needs inverse of eMat, 
which is a problem if Det[eMat]=0*)
getHHNRotPEmat [eMat_] :=  Module[{rootEtE, rotInv, rotOpt}, 
       rotInv = Inverse[eMat];
       rootEtE = MatrixPower[Transpose[eMat] . eMat, +1/2];
       rotOpt = rootEtE . rotInv]

(* 5  HNN "Minus root" Our Preferred version does not need inverse of eMat,  
     just inverse of the S factor and a Transpose.*)
getHHNRotMEmat [eMat_] :=  Module[{invRootEtE, rotOpt}, 
       invRootEtE = MatrixPower[Transpose[eMat] . eMat, -1/2];
       rotOpt = invRootEtE . Transpose[eMat]]
             
  }  

\end{document}